\documentclass[runningheads]{llncs}

\usepackage{eccv}

\usepackage{eccvabbrv}

\usepackage{graphicx}
\usepackage{booktabs}

\usepackage[accsupp]{axessibility}  

\usepackage{hyperref}

\usepackage{orcidlink}

\usepackage[table]{xcolor}
\usepackage{tabularx}
\usepackage{float}
\usepackage{mfirstuc}
\usepackage{multirow}
\usepackage{multicol}
\usepackage{bbding}
\usepackage{xstring}
\usepackage{algorithm}
\usepackage[noend]{algpseudocode}
\usepackage{tcolorbox}
\tcbuselibrary{skins,breakable,listings}
\usepackage{fancyvrb}
\usepackage{fvextra}
\usepackage{longtable}

\newcommand{\cha}{\XSolidBrush}
\newcommand{\gou}{\CheckmarkBold}

\newcommand{\ddelta}[1]{\textcolor{purple}{#1}}
\newcommand{\cdelta}[1]{\textcolor{ForestGreen}{+#1}}

\begin{document}

\title{Generalize LMMs to Versatile Visual Modalities via Fabricated Modality Synthesis}

\titlerunning{Generalize LMMs to Versatile Visual Modalities}

\author{ 
Shihao Yuan 
\inst{1}
\orcidlink{0000-0002-2022-6774}
\and 
Yuanze Li
\inst{1}
\orcidlink{0009-0009-7365-1158}
\and
Ruyi Zhang
\inst{1}
\and\\
Ming Liu
\inst{1}$^($\Envelope$^)$
\orcidlink{0000-0001-9136-8481}
\and
Wangmeng Zuo
\inst{1}
\orcidlink{0000-0002-3330-783X}\\[0.5em]
\texttt{\{csshihao, sqleopop, csmliu\}@outlook.com},\\\texttt{\{ruyi.zhang.maggie, cswmzuo\}@gmail.com}
}

\authorrunning{Shihao Yuan, Yuanze Li,~\etal}

\institute{
$^1$Faculty of Computing, Harbin Institute of Technology, Harbin, China
}

\maketitle

\begin{abstract}
Despite the advancements of Large Multimodal Models~(LMMs) in RGB vision, their ability to generalize to unseen visual modalities remains a largely unexplored challenge.
We argue that different visual modalities are merely distinct samplings of the same physical world. 
Therefore, effective generalization requires models to possess both modality-agnostic perception of scene semantics and the adaptability to modality-specific characteristics. 
To achieve this, we propose a training framework, \textbf{VVM-Tuning}, to equip LMMs with these capabilities through modality synthesis and modality contexts. 
Specifically, we synthesize diverse appearance-varied images from RGB scenes, training the model to disentangle invariant semantics from varying visual appearances, and align these appearances with language for visual concepts decoupled from modalities.
We then introduce modality contexts in the prompt and use instruction tuning to assist the model in mapping these appearance variations back to modality-related attributes, enabling zero-shot adaptation to unseen modalities during inference. 
To facilitate research in this direction, we introduce \textbf{VVM-Bench}, a comprehensive benchmark featuring 6 real and synthetic modalities to evaluate semantic perception and modality understanding. 
Experiments demonstrate that, via our training on synthetic modalities, 5 tested models exhibit consistent improvements on both real-world and novel synthetic modalities without in-modality training. 
Source code and data will be publicly available at \url{https://github.com/Hunter-Will/VVM-Tuning}.

\end{abstract}    
\section{Introduction}
\label{sec:intro}
\begin{figure}[ht]
    \centering
    \includegraphics[width=0.95\textwidth]{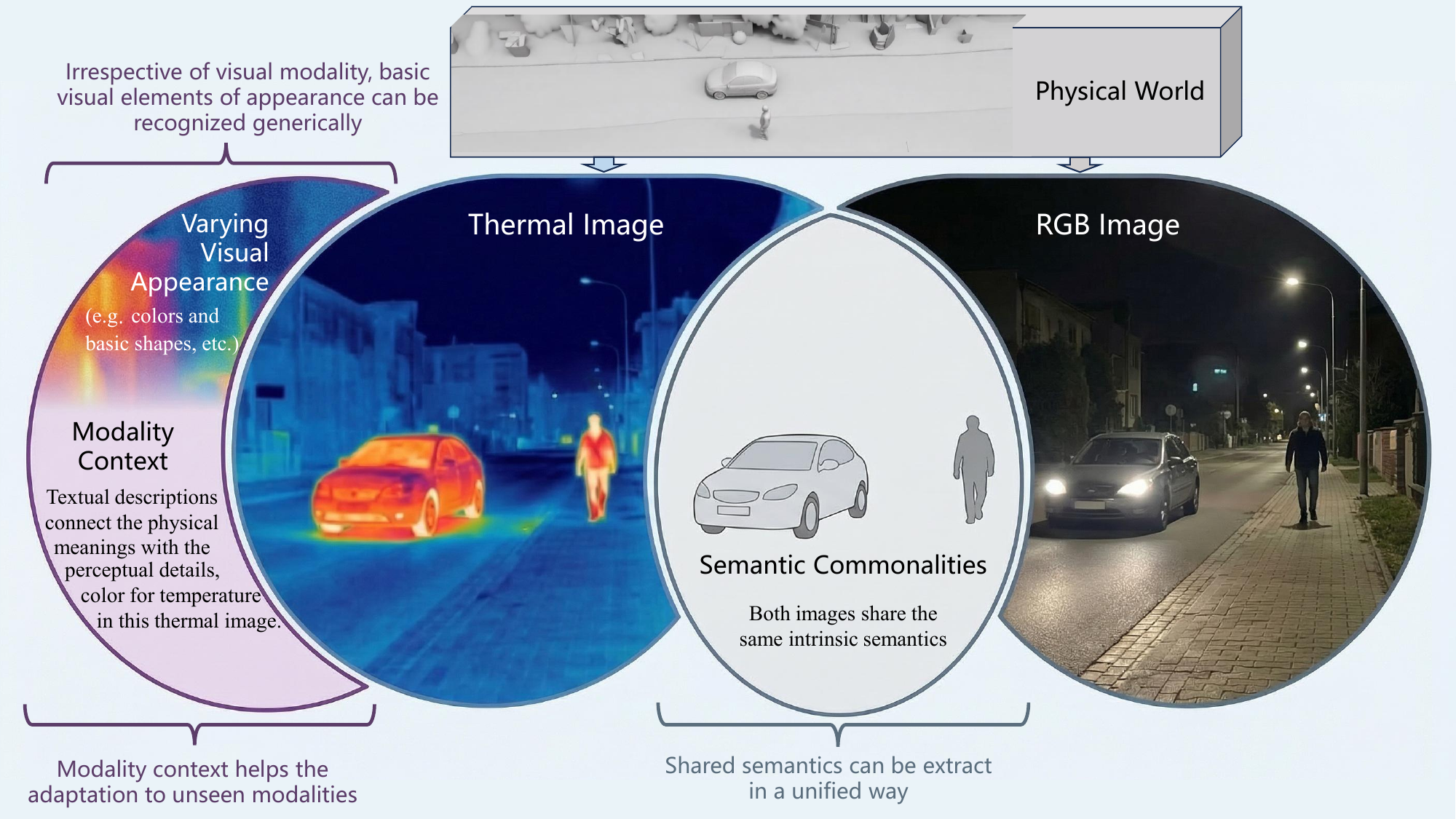}
    \caption{An overview of our idea.  Different visual modalities~(Thermal and RGB) are both signal samplings of our physical world, sharing the same underlying semantics~(formed by the person and the car). They also share basic visual concepts, as the color itself is invariant across modalities. The modality-specific physical meaning is encoded by the rearrangement and remapping of basic visual concepts. Thus, we introduce textual context to complement such knowledge during inference.}
    \label{fig: 1_teaser}
    \vspace{-0.2in}
\end{figure}

Current Large Multimodal Models~(LMMs) demonstrate impressive visual performance on RGB images\cite{mme_survey}; however, the generalization boundary on other visual modalities~(\eg infrared, depth, \etc) remains largely unexplored yet. 
Existing non-RGB vision in current LMMs primarily depends on the non-RGB data incorporated during the data scaling process\cite{qwen3vl,llavaov15,internvl_35}. 
This leads to a generalization gap when encountering unseen modalities, as the non-RGB vision is built on in-modality data and limited modality coverage.

To address this gap and explore the generalization boundary of LMMs, we raise the question: 
\textit{Is it possible for LMMs to generalize across versatile visual modalities without training on in-modality data?}

\begin{figure}[t]
    \centering
    \includegraphics[trim=0cm 0cm 0cm 0cm,clip=true,width=0.8\textwidth]{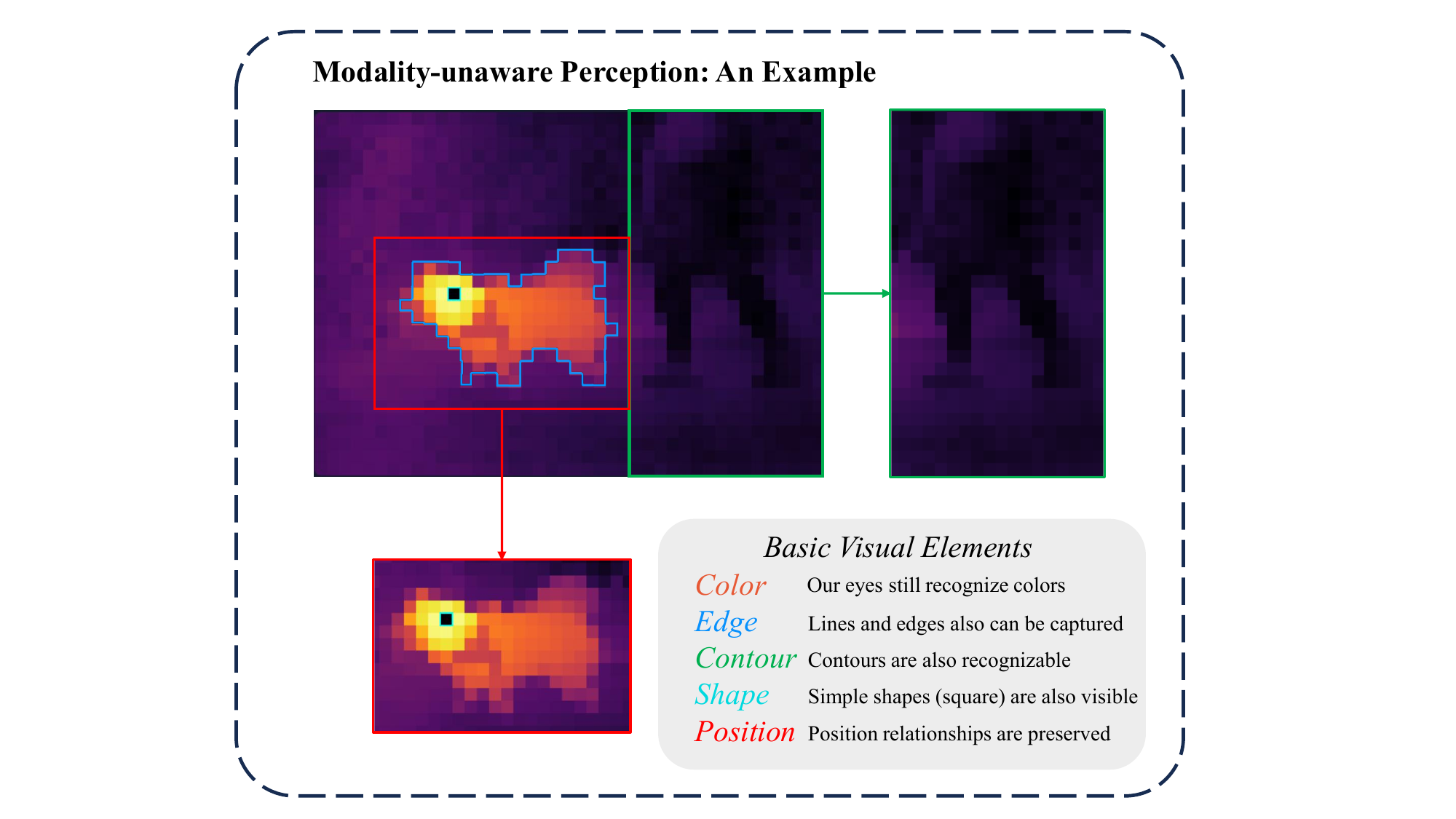}
    \vspace{-0.1in}
    \caption{An example of how human vision adapts to unseen visual modalities. Our eyes can capture basic visual elements regardless of modalities, and our brain recognizes the semantics even through a few visual elements. This demonstrates a compatible modality-unaware perception, which we could implement in LMMs' training.}
    \label{fig: 1_example}
    \vspace{-0.2in}
\end{figure}

The solution to the problem starts from the nature of visual modalities, as shown in \cref{fig: 1_teaser},  though with different appearances, they are essentially digital signals collected by sensors~(thermal and RGB camera in \cref{fig: 1_teaser}) from the same physical space. Therefore, images from different visual modalities can be viewed as distinct samplings of identical real-world scenes. 
This commonality lies the foundation for pan-modal generalization, as the understanding of underlying semantics is modality-agnostic.
Furthermore, through observation, we find that in human vision, images from other visual modalities can still be \textbf{seen} even without understanding the modality. Take \cref{fig: 1_example} as an example, our brain can still read the semantics of the image from basic visual elements. This gives us inspiration that LMMs can learn this process to recognize semantics even without certain modality knowledge, which we call Modality-unaware Perception. 
Also, those visual elements, belonging to general visual concepts, are not completely combined with semantics. Some of them~(like the colors in the thermal image in \cref{fig: 1_teaser}) might not be a necessary part for understanding the whole image semantics under this modality-unaware circumstance.
Instead, they are supposed to combine with the physical meaning of the visual modality to achieve a higher understanding of the image, such as the color-coded thermal image in \cref{fig: 1_teaser}.
Although they can only be understood superficially without modality knowledge, they are still basic elements of both human vision and machine vision, which are completely generalized among diverse visual modalities, as long as they are 3-channel digital images. 
That means they are still perceivable modality-unawarely, for example, the thermal color in \cref{fig: 1_teaser} can still be understood as colors without knowing its connection with temperature.
We refer to this part that only perceives basic visual elements without understanding further as \textit{photographic perception}. Correspondingly, the part that comprehends underlying semantics is referred to as \textit{semantic perception}. Both parts compose a modality-unaware perception without modality knowledge.
This modality-unaware perception demonstrates a highly possible approach of unseen visual modality generalization; however, it doesn't mean that every single visual modality can be understood in the same way. 
Different visual modalities have their unique sampling function of the real world, which is defined by the sensor and signal processing procedure. This often leads to varying appearances of the same underlying semantics and usually means a connection between the physical attributes of entities in the real-world scene and the unique visual appearances in digital images, which is, of course, different from regular RGB camera images. Besides, non-RGB modalities usually map an additional physical quantity to the color of the image, known as the \textit{pseudo-color}\cite{pseudo_1975,pseudo_2012}, which actually enriches the meaning of basic visual elements. 
Understanding such a connection or mapping requires modality knowledge, and such modality knowledge is not reachable during training for the unseen visual modalities in our discussion. Thus, we introduce modality contexts to complement the modality understanding by utilizing the instruction following and in-context learning ability of LMMs during inference.

Viewing LMMs' training from this perspective, we argue that the current LMMs are largely overlooking the modality-unaware perception. The visual-language alignment, building primarily on RGB images, doesn't distinguish the invariant semantics and varying visual appearances. This would cause an over-alignment and isolate other visual modalities, as the model needs to learn new alignments for each different visual modality incorporated in training.
Based on the above insights, we proposed a novel training framework called \textbf{VVM-tuning} to: 
1) Build up a modality-unaware perception via fabricated images;
2) Enable the modality-aware understanding of LMMs through modality synthesis.

For the first part, we leverage RGB images to fabricate non-RGB images to mimic varying appearances from different visual modalities, keeping the semantics invariant from RGB images. Therefore, the model has a chance to learn how to read invariant semantics through varying appearances. 
Furthermore, we design two sets of descriptions and VQA tasks to disentangle the semantic parts and perceptual parts of the visual-language alignments. 
By separate alignment, the model is taught to extract the semantics and keep perceptual elements aligned with language at the same time, which would benefit the modality-aware understanding for reconnecting the physical meanings and visual appearances.
Because of the fabricated image, we can largely diversify the visual appearances of common scenes; thus, we establish a more general perception of visual appearances, which leads to a base for modality-aware understanding.

For the second part, we utilize modality synthesis to build a general synthetic instruction set to train the model with modality contexts.
These modality contexts are fabricated to construct diverse synthesized visual modalities combined with the fabricated images. 
From these synthetic data, the model learns to connect the modality-specific physical attributes to varying visual appearances and further integrates with the semantics to achieve better modality understanding. 
With the combination of the above two parts, we enable the zero-shot generalization of LMMs to unseen visual modalities without in-modality data for training.

Finally, we assemble a diverse visual modality benchmark, VVM-bench, containing 6 real and synthesized imaging modalities.
From our perspective, this benchmark evaluates the generalization ability through 11 VQA tasks from modality-unaware perception and modality-aware understanding, respectively. Half of the 6 non-RGB visual modalities are collected from VS-TDX\cite{mspr}, and 3 fabricated modalities to mimic unseen visual modalities. 
Experiments show an average of 8.2\% improvements of perception tasks and 3.8\% improvements of understanding tasks on this benchmark among 5 different base models and in 6 synthesized and real visual modalities for both perception and understanding. 

In summary, our contribution can be listed as follows:
\begin{itemize}
\item To our best knowledge, we are the first to explore the potential of LMMs to generalize to unseen visual modalities, and demonstrate a possible approach through modality context and modality synthesis. 

\item We introduce synthetic data for other visual modalities and propose a fabricated modality synthesis data pipeline, which can produce diverse fabricated images and synthesized modality contexts.

\item We propose a training framework establishing modality-unaware perception and modality-aware understanding, and confirm that LMMs can generalize to other visual modalities via modality synthesis and disentangled tasks.

\item We assemble a versatile visual modality benchmark to evaluate the visual ability and generalization on visual modalities for LMMs.
\end{itemize}
\section{Related Works}
\label{sec:related}
Since the emergence of LLaVA\cite{llava}, establishing the visual instruction tuning paradigm, both LMMs and benchmarks targeting their RGB visual capabilities have seen significant development\cite{mme_survey}. However, the generalization boundary of LMMs to non-RGB visual modalities remains underexplored. 
Modern LMMs, such as LLaVA-OneVision\cite{llavaov,llavaov15}, Qwen-VL\cite{qwenvl,qwen25vl,qwen3vl}, and InternVL\cite{internvl,internvl3,internvl_35}, incorporate diverse and abundant visual data, which may contain other modalities in their training subsets like DocQA\cite{docvqa} and ScienceQA\cite{scienceqa}. However, their primary focus remains on scaling data and diversity to improve general visual abilities, rather than investigating the generalization of unseen visual modality.

Meanwhile, some previous works, such as Imagebind\cite{imagebind} and PandaGPT\cite{pandagpt}, also involve thermal and depth images. However, the basic idea of Imagebind\cite{imagebind} is “binding” other general modalities~(audio, text, \etc.) to RGB images in order to eliminate the need for pairwise alignments between these modalities. As a result, RGB-D and RGB-T paired data are still required during the training of Imagebind\cite{imagebind}, which are not regarded as \textbf{unseen} visual modality and not within our scope of discussion. Similarly, the cross-modal capabilities discussed in PandaGPT\cite{pandagpt} refer to the zero-shot instruction following ability among 6 seen modalities inherited from Imagebind\cite{imagebind}, which is also beyond the discussion.
Concurrently, research dedicated to non-RGB modalities has largely been confined to fine-tuning LMMs on specialized datasets. For instance, Infrared-LLaVA\cite{infraredllava} adapts LLaVA for infrared-specific tasks by tailoring the model to the infrared modality.
SpatialBot~\cite{spatialbot} focuses on the spatial understanding with both RGB and depth images, also relying on RGB-D training to inject depth knowledge. Specialized LMMs, for instance, EarthGPT\cite{earthgpt} in remote sensing, XrayGPT\cite{xraygpt} and RadVLM\cite{radvlm} in radiology, are trained with domain-specific data. Though they also involve multiple visual modalities, they are focusing on application tasks, leaving their inherent ability to generalize across different useen visual modalities behind. 

This gap is further reflected in existing evaluation suites. Mainstream benchmarks for general-purpose LMMs, such as MMBench~\cite{mmbench}, MME~\cite{mme}, MMMU~\cite{mmmu}, MM-Vet~\cite{mmvet}, and SEED-Bench~\cite{seedbench}, are predominantly centered on RGB images. While a few recent efforts have extended evaluation to non-RGB modalities for LMMs, the RGB-Th-Bench~\cite{rgbtbench} focuses on the cross-reference understanding of RGB-T paired images, ignoring the possible generalization to unseen modalities on general visual abilities. The VS-TDX~~\cite{mspr} benchmark evaluates general visual abilities on 3 modalities, but still lacks focus on the principles and testing of unseen visual modality generalization. Other specialized evaluations, like MedSG-Bench\cite{MedSG-Bench} in medical imaging and RSVQA\cite{rsvqa} in remote sensing, are highly domain-specific and not designed to assess general visual adaptability. Consequently, there remains a lack of benchmarks tailored to systematically evaluate the cross-visual-modal generalization capabilities of LMMs. 
These limitations underscore a critical gap in understanding the true visual versatility of modern LMMs beyond the RGB vision, motivating our study to investigate their generalization potential across different visual modalities.

\section{Methods}

\subsection{Preliminary}
From the base idea of how different visual modalities follow the basic imaging principles to map the real-world scenarios into photos, our method tries to imitate the same process of how humans understand versatile visual modalities, which leads to two different abilities: modality-unaware perception and modality-aware understanding.
To equip LMMs with these two abilities, we design two sets of generated instruction tasks\cite{instructiongen} and train the model to learn the adaptability from these tasks with fabricated images and modalities. Our VVM-Tuning mainly follows the paradigm of visual instruction tuning\cite{llava}.
The training objective follows the common approach of Supervised Fine-Tuning~(SFT), which involves the Cross Entropy loss function $\mathcal{L}_{SFT}$. 
\begin{align}
\label{eq: l_sft}
\mathcal{L}_{SFT}&=-\sum_{i=1}^{n}{\log \pi_{\theta}(y_i|X,y_1,y_2,...,y_{i-1})},
\end{align}
The whole formalized expression of the training is to optimize $\mathcal{L}_{SFT}(T_{a}, T_{gt})$ to minimize the differences between the generated $T_{a}$ and the ground truth $T_{gt}$.

\subsection{Modality Synthesis}
\label{sec: Modality Synthesis}
\begin{figure}[t]
    \centering
    \includegraphics[trim=0cm 1cm 0cm 0.5cm,clip=true,width=0.95\textwidth]{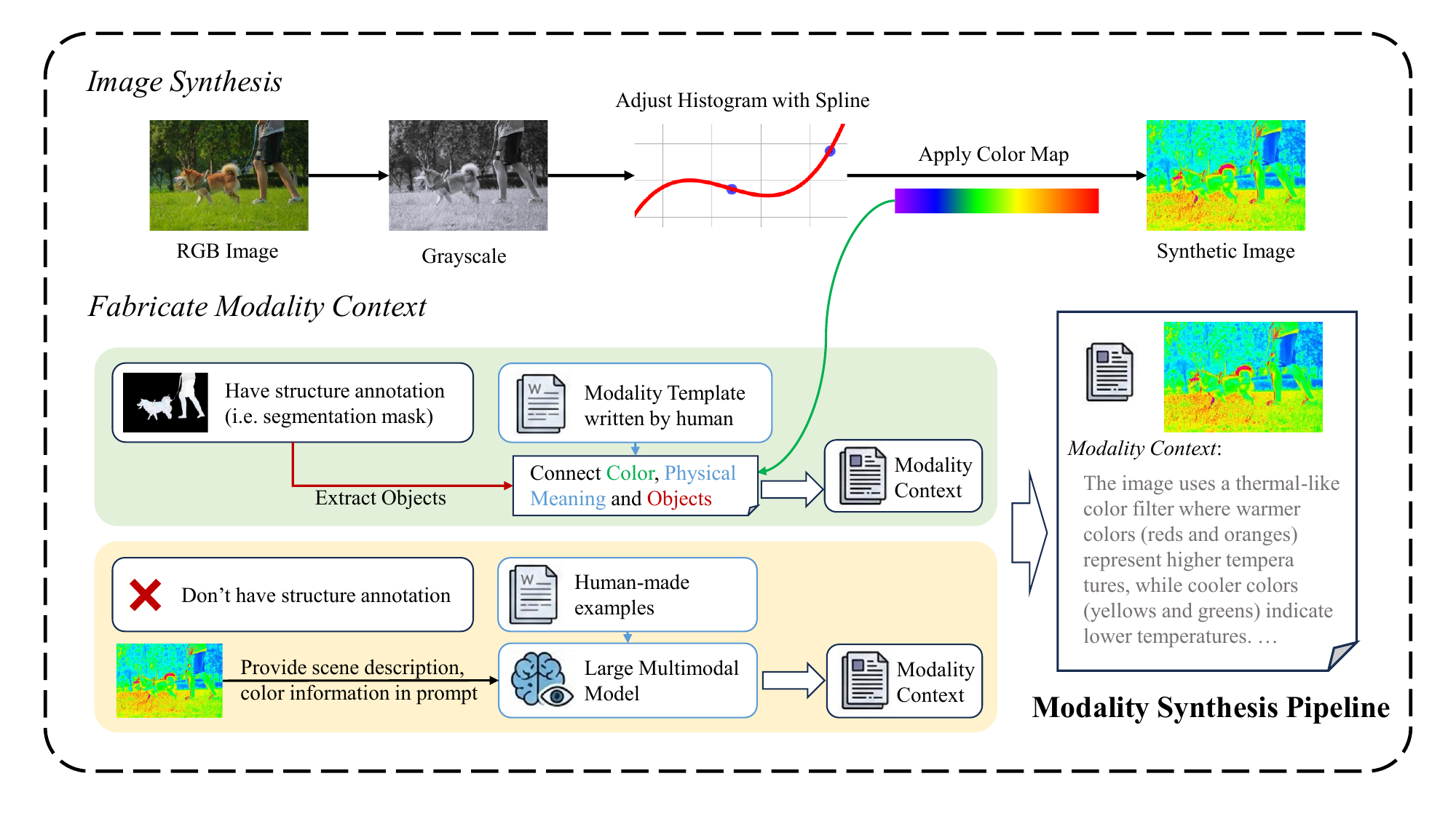}
    \caption{An overview of our modality synthesis pipeline, composed of the image synthesis and modality context fabrication. 
    }
    \label{fig: syn_pipe}
    \vspace{-0.15in}
\end{figure}

Effective generalization requires cross-modal perception and context-guided modality adaptation, which needs the training data covering versatile modal characteristics. Thus, we introduce a visual modality synthesis pipeline, shown in \cref{fig: syn_pipe}. The pipeline is mainly composed of two parts: 1) image synthesis and 2) modality context fabrication.

\begin{figure}[b]
    \vspace{-0.2in}
    \centering
    \includegraphics[trim=0cm 5cm 0cm 5cm, clip=true,width=0.95\textwidth]{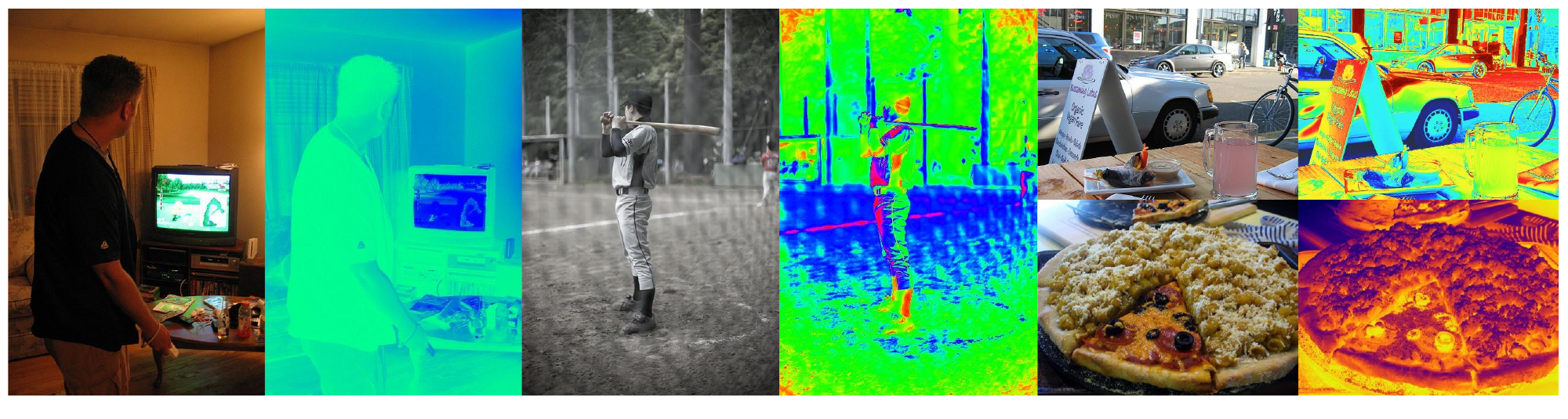}
    \vspace{-0.1in}
    \caption{Different examples of fabricated images with different color maps produced by our image synthesis pipeline. 
    }
    \label{fig: syn_example}
\end{figure}

The first part is to imitate the unique characteristics that differ from regular RGB photos, such as the pseudo-color pattern\cite{pseudo_infrared} and histogram distribution. Here, we introduce a simple pipeline to fabricate non-RGB images from RGB images.   
As shown in \cref{fig: syn_pipe}, the RGB source images are first converted to grayscale for histogram adjustments. We apply cubic spline interpolation\cite{spline} to randomize the shape of the histogram\cite{splines_hist} and increase the diversity by flipping the whole histogram randomly. Then the grayscale images are painted with pseudo-color by randomly applying 22 color maps\cite{colormaps} that are commonly used in scientific computing and visualization. 
After color mapping, several image augmentations\cite{imageaug}, such as noise injection, smoothing, and blurring, are employed to simulate the noisy and low-resolution modalities. With such a simple pipeline, we manage to fabricate diverse images with similar characteristics to the real non-RGB visual modalities, shown by the examples in \cref{fig: syn_example}.

Besides fabricated images, we also make up corresponding physical meanings and explanations to compose modality contexts. This is done by two approaches according to whether there are structure annotations~(segmentation mask or object information) of source RGB images, shown in \cref{fig: syn_pipe}. For the images with annotations, the modality contexts are assembled structure texts by connecting the color maps with objects in the image by labels such as segmentation masks. Then, the value interval mapping colors is ascribed with human-made physical attributes and fit into manual templates to form modality contexts.
For the non-annotated images, we leverage the in-context learning ability of LMM to generate color and object-related modality context by prompting it to follow manually written examples. 
Combining fabricated images and synthesized modality contexts results in modality synthesis that contains diverse characteristics to teach LMMs the varying and invariant features of versatile visual modalities. Further details of the modality synthesis pipeline can be found in the Appendix.

\begin{figure}
    \centering
    \includegraphics[width=0.95\textwidth]{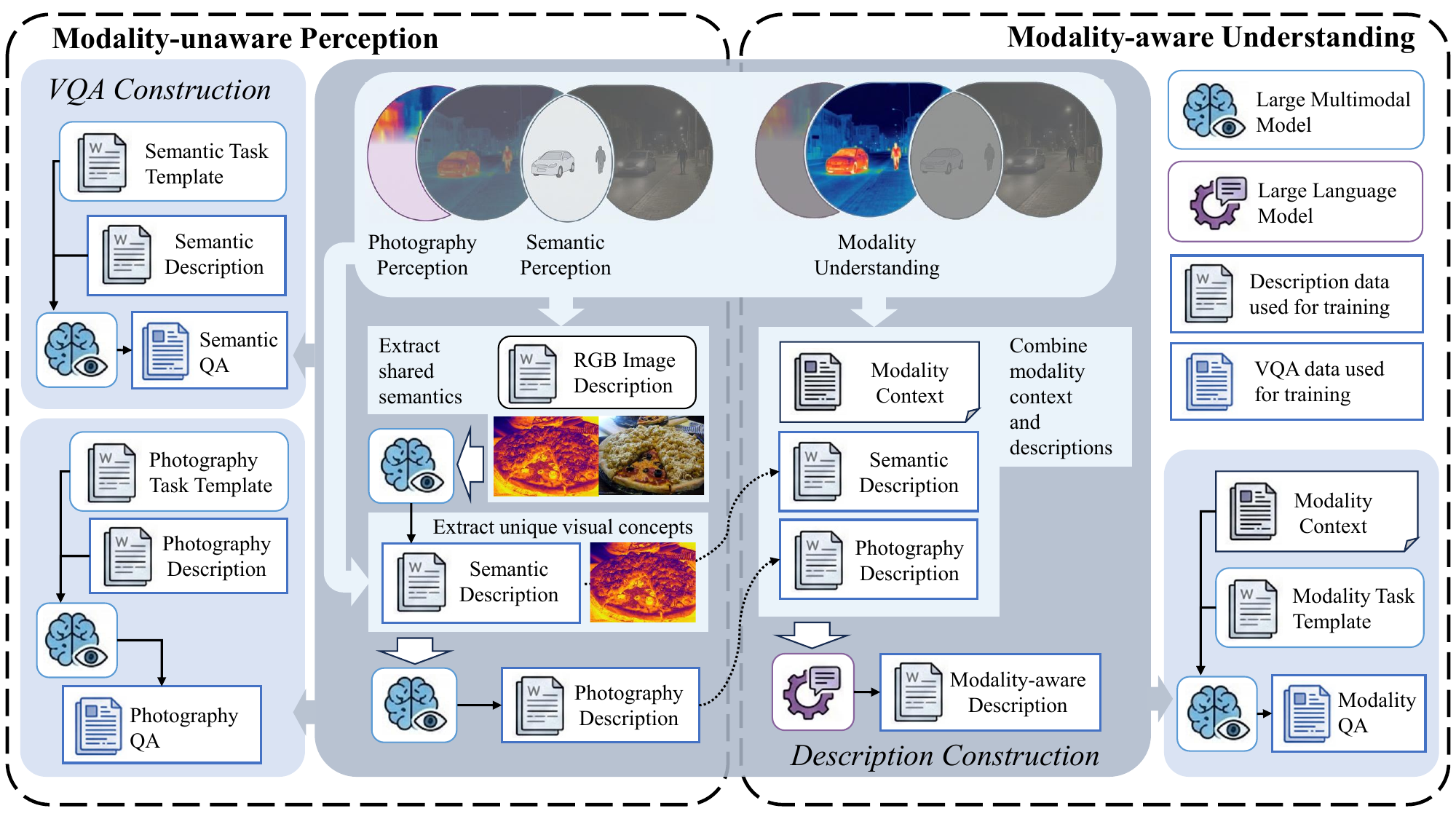}
    \caption{An overview of the task construction of \textbf{VVM-Tuning}, including modality-unaware perception tasks and modality-aware understanding tasks. Starting from our core insight at the top, the region illustrates the construction of the semantic description and photographic description tasks to form modality-aware descriptions. Then, the surrounding VQA tasks are generated from task templates with descriptions.
    }
    \label{fig: task_overview}
    \vspace{-0.2in}
\end{figure}

\subsection{Modality-unaware Perception}

The foundation of generalizing to unseen visual modalities is an invariant perception ability across them. This ability is called modality-unaware perception, which contains two parts: 1) the underlying semantic perception and 2) the photographic perception of basic visual elements. 
In regular RGB visual-language alignment, these two parts of perception are usually entangled together, aligned with the language as a whole. However, when other visual modalities are involved, the entangled perception isolates the shared semantics between visual modalities due to different appearances. This causes the LMMs need to learn new alignments for each different visual modality. 
As the underlying semantics are encoded in different forms of expression by different visual modalities, we first design scene semantic tasks to equip LMM with the ability to extract the underlying semantics. Then, photographic perception tasks are involved to capture unique visual clues introduced by different visual modalities. 
With the image synthesis pipeline, the fabricated images are convenient to create the same semantics with different visual appearances. Based on them, we design several instructions to train the model to recognize the physical entities in various appearances and activate the common knowledge with such entities.
Furthermore, the photographic perception tasks ensure the LMM captures the visual details and remains loyal to the image instead of being confused by semantics.

For better adaptability, both semantic perception and photographic perception are composed of two forms of tasks, image description\cite{description} and visual question-answering~(VQA)\cite{vqa,llava}. As shown in the left part of \cref{fig: task_overview}, they form four tasks: semantic description, semantic QA, photographic description, and photographic QA. 
First, for the semantic description, we leverage the common semantics of fabricated images and prompt an LMM to extract the overlapped parts from the description of the RGB image according to the RGB image and the corresponding fabricated image. That leads to a semantic description that extracts the shared semantics of fabricated and RGB images. Then, the photographic description is generated by prompting the LMM to extract the unique parts from the fabricated image that are not in the semantic description. However, due to the lack of perception of these fabricated images, the colors in the photographic description need to be recorrected by introducing the color names during the image synthesis pipeline. As for the VQA instructions, we design 11 different sub-tasks for semantic and photographic QA as human-made question templates and utilize a large-sized LMM to generate diverse questions and answers according to RGB images, fabricated images, and corresponding descriptions, respectively. Finally, the questions are randomly organized as a mix of non-choice questions and choice questions to combine with the descriptions to create a completely modality-unaware training set. Further details of the task distribution and templates can be found in the Appendix.

\subsection{Modality-aware Understanding}

The purpose of this part of VVM-Tuning is to enhance the understanding of modality contexts.
Modality context is a textual prompt, including specific characteristics and physical meanings of the modality of the image. The two ways of synthesizing modality context are described in \cref{sec: Modality Synthesis}. The modality-aware understanding tasks leverage synthesized modality contexts to equip the model with the ability to understand modality contexts and connect the physical meaning to specific visual concepts or appearances. Same as the modality-unaware perception, the modality-aware training also consists of image description and VQA tasks, called modality description and modality QA, shown in the right part of \cref{fig: task_overview}. The construction of modality description is achieved by leveraging a large language model to combine the semantic description, photographic description, and synthesized modality context together to form an image description with modality knowledge. 
The modality QA is generated by LMM according to the modality context and human-made templates. We don't use the modality description here as the VQA tasks are concentrated to particular part of the image and the modality description only provides a general description of the entire image. However, scene and photographic descriptions are involved to make sure the LMM for generation understands the non-RGB images enough. The question templates of modality QA are specially designed to mainly cover different circumstances that require a knowledge of the visual modality. Further details can be found in the Appendix.

\subsection{Benchmark}

\begin{figure}[htb]
    \centering
    \includegraphics[width=0.9\textwidth]{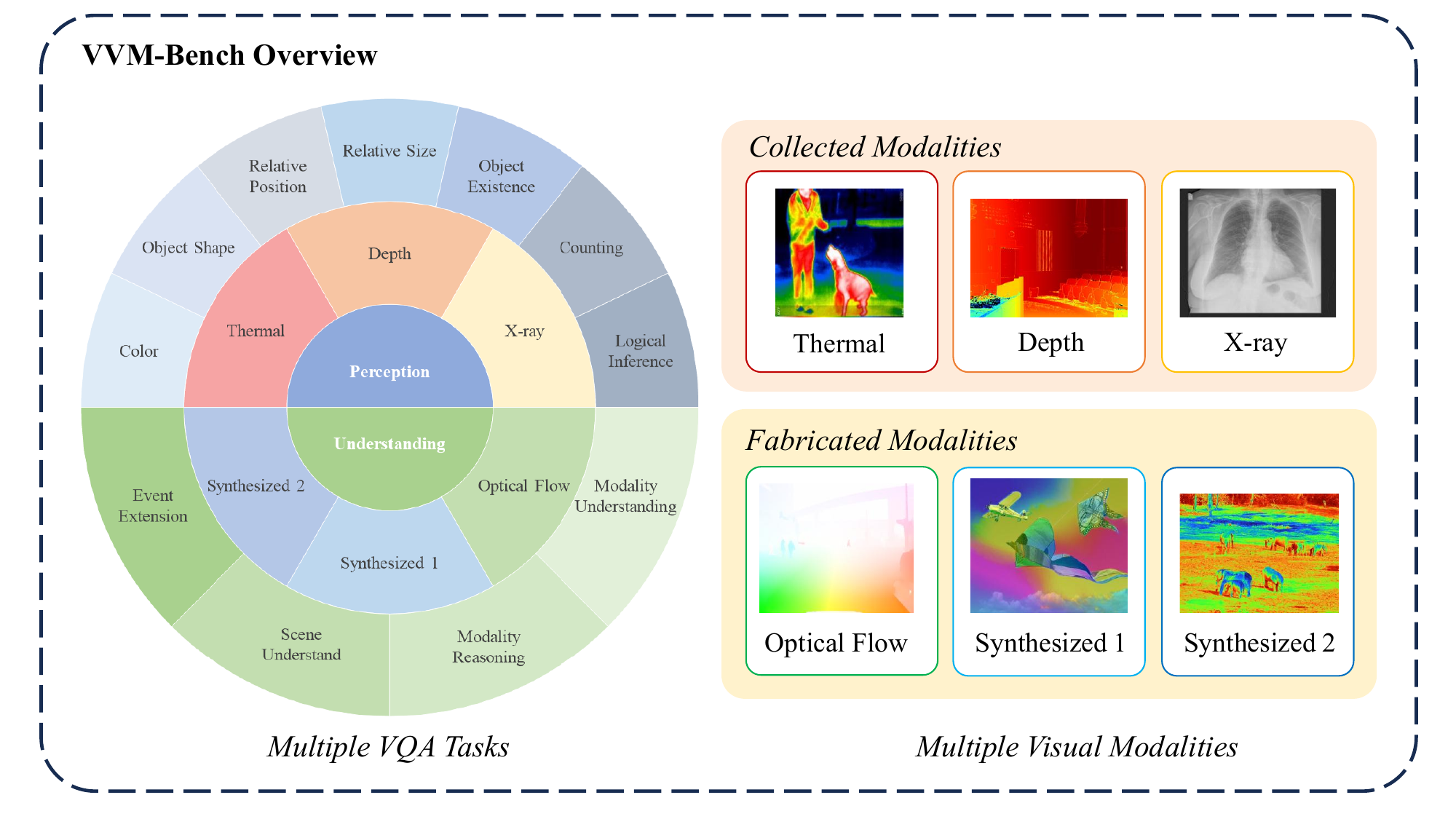}
    \vspace{-0.1in}
    \caption{An overview of VVM-Bench. Our VVM-Bench contains 6 visual modalities and 11 VQA tasks for perception and understanding evaluation, including 3 real modalities collected from previous work and 3 fabricated from RGB images.}
    \label{fig: bench_overview}
    \vspace{-0.2in}
\end{figure}

To mitigate the lack of evaluation data and confirm the feasibility of our training framework, we assembled a benchmark that contains multiple real and synthesized visual modalities to test the perception and understanding abilities of LMMs. The VVM-Bench, as shown in \cref{fig: bench_overview}, is composed of 6 visual modalities and 11 VQA tasks belonging to perception and understanding, respectively. 
Among them, three modalities~(thermal, depth, and X-ray) are collected from VS-TDX\cite{mspr} with minimal changes of output formation to insert modality contexts in understanding tasks. The other three modalities are fabricated from RGB images, including Optical Flow, Synthesized 1, and Synthesized 2. 
Specifically, the optical flow images are visualizations of predicted optical flows from RAFT\cite{raft} to serve as signals from an almost entirely new and unknown imaging sensor with real physical meanings.
The other two visual modalities are completely synthesized as unseen visual modalities. The images in Synthesized 1 are fabricated using a style transfer algorithm, StyleShot\cite{styleshot}, to apply a visualization style on normal RGB images.
As for images for Synthesized 2, we use the same image synthesis method as described in \cref{sec: Modality Synthesis} with changed hyperparameters to explore the generative diversity of our image synthesis pipeline.

Our benchmark evaluates two different aspects of LMMs corresponding to our insights, the modality-unaware perception and modality-aware understanding, abbreviated as perception and understanding in \cref{fig: bench_overview}. 
The perception aspect contains 7 sub-tasks, and all tasks focus on the semantics or visual attributes perceivable without modality knowledge. No modality contexts are provided in these tasks.
Instead, the understanding tasks are all about the modality understanding or knowledge association with the visual modality, which provides modality contexts in the prompts.
All modality contexts are written by human and we use choice questions in VVM-bench for the convenience of evaluation.
Detailed task templates and modality contexts can be found in the Appendix.

\section{Experiments}

\subsection{Implementation Details}
We have employed the training framework on multiple base LMMs, including LLaVA-1.5\cite{llava}, Qwen2.5-VL\cite{qwen25vl}, and Qwen3-VL\cite{qwen3vl} with different sizes in our capabilities. All the experiments are conducted on a 4 NVIDIA A800 GPU server with a total batch size of 32. Following the common setting of Supervised Fine-Tuning~(SFT), we apply the AdamW\cite{adamw} as optimizer and set the learning rate to $1e-5$. 
We adopt a single-staged fine-tuning with roughly 50k training samples in total for 1 epoch, involving all tasks in modality-unaware perception and modality-aware understanding. To preserve the general abilities of LMMs, we retain 12k RGB visual instruction data in the 50k training dataset.

Our benchmark follows the common VQA benchmark setting, with the form of choice questions. To quantify the results, the LMMs are instructed to output the option's letter only to calculate accuracy. Specifically, for tasks that need modality contexts, all modality contexts are inserted between the image and the question as contexts. Further evaluation details can be found in the Appendix.

\subsection{Results of Modality-unaware Perception}
The results in \cref{tab: vvm-perce} show the VQA accuracy on the perception subset of VVM-Bench. The results in \cref{tab: vvm-perce} report the total accuracy of all samples in each modality for both baselines and our tuned models.
From the results in \cref{tab: vvm-perce}, experiments show our VVM-tuning brought an average 8.2\% improvement for all base models among 6 modalities on our VVM-Bench perception subset. 
Across all tested models, LLaVA-1.5 gained the biggest average improvement of 13.6\% because of its relatively low base. The 4B model of Qwen-3-VL got the lowest improvement, but still exceeded 5\%. 
This exhibits the adaptability of our modality-unaware perception training among different base models and confirms that the perception of LMMs on unseen visual modalities still has room for improvement.
\begin{table}[h]
\vspace{-0.2 in}
\centering
\caption{The results of the Modality-unaware Perception subset of VVM-bench. “$\Delta$” represents accuracy difference between the model we tuned and the baselines. “Average” shows the average accuracy and difference across all tested modalities. “Average $\Delta$” shows the average improvements across all tested models for each visual modality. Note that there is no modality context for this perception subset.}
\label{tab: vvm-perce}
\resizebox{\textwidth}{!}{\begin{tabular}{c|c|cccccc|c}
\toprule
\multirow{2}{*}{Model}& \multirow{2}{*}{Size} & \multicolumn{3}{c}{Collected Visual Modalities} & \multicolumn{3}{c|}{Fabricated Visual Modalities} 
& \multirow{2}{*}{Average}\\
& & Thermal & Depth & X-Ray
& Optical Flow & Synthesized 1 & Synthesized 2 & \\ \midrule
LLaVA-1.5 & 7B & 55.6\% & 62.4\% & 58.1\% & 53.2\% & 70.7\% & 62.6\% & 60.4\% \\
LLaVA-1.5~(Ours) & 7B & 69.1\% & 69.8\% & 67.3\% & 75.4\% & 85.8\% & 77.0\% & 74.1\% \\
$\Delta$ & - & \cdelta{13.5\%} & \cdelta{7.4\%} & \cdelta{9.2\%} & \cdelta{22.2\%} & \cdelta{15.1\%} & \cdelta{14.4\%} & \cdelta{13.6\%} \\
Qwen-2.5-VL & 3B & 72.0\% & 67.9\% & 68.1\% & 67.2\% & 83.0\% & 74.9\% & 72.2\% \\
Qwen-2.5-VL~(Ours) & 3B & 81.7\% & 77.1\% & 77.5\% & 83.1\% & 88.9\% & 83.0\% & 81.9\% \\
$\Delta$ & - & \cdelta{9.7\%} & \cdelta{9.2\%} & \cdelta{9.4\%} & \cdelta{15.9\%} & \cdelta{5.9\%} & \cdelta{8.1\%} & \cdelta{9.7\%} \\
Qwen-2.5-VL & 7B & 74.5\% & 73.3\% & 75.7\% & 77.0\% & 87.1\% & 81.1\% & 78.1\% \\
Qwen-2.5-VL~(Ours) & 7B & 79.8\% & 79.2\% & 79.6\% & 88.4\% & 91.2\% & 86.4\% & 84.1\% \\
$\Delta$ & - & \cdelta{5.3\%} & \cdelta{5.9\%} & \cdelta{3.9\%} & \cdelta{11.4\%} & \cdelta{4.1\%} & \cdelta{5.3\%} & \cdelta{6.0\%} \\
Qwen-3-VL & 4B & 80.6\% & 77.8\% & 76.7\% & 73.0\% & 81.6\% & 75.6\% & 77.6\% \\
Qwen-3-VL~(Ours) & 4B & 82.6\% & 84.4\% & 79.5\% & 81.5\% & 86.9\% & 82.1\% & 82.8\% \\
$\Delta$ & - & \cdelta{2.0\%} & \cdelta{6.6\%} & \cdelta{2.8\%} & \cdelta{8.5\%} & \cdelta{5.3\%} & \cdelta{6.5\%} & \cdelta{5.3\%} \\
Qwen-3-VL & 8B & 79.8\% & 78.5\% & 73.8\% & 73.3\% & 82.4\% & 78.0\% & 77.6\% \\
Qwen-3-VL~(Ours) & 8B & 82.1\% & 84.2\% & 80.3\% & 86.5\% & 89.5\% & 82.1\% & 84.1\% \\
$\Delta$ & - & \cdelta{2.3\%} & \cdelta{5.7\%} & \cdelta{6.5\%} & \cdelta{13.2\%} & \cdelta{7.1\%} & \cdelta{4.1\%} & \cdelta{6.5\%} \\ \midrule
Average $\Delta$ & - & \cdelta{6.6\%} & \cdelta{7.0\%} & \cdelta{6.4\%} & \cdelta{14.2\%} & \cdelta{7.5\%} & \cdelta{7.7\%} & \cdelta{8.2\%} \\
\bottomrule
\end{tabular}}
\end{table}

Meanwhile, although training on synthetic data, all base models show at least an average 6\% improvement on visual modalities with real physical meanings, as shown in Thermal, Depth, X-ray, and Optical Flow. 
This confirms the existence of the generalization foundation, and different real visual modalities do share commonalities in both semantics and perceptual elements.
Also, the performance on synthesized modalities indicates that this improvement could generalize to completely unseen visual modalities and further confirm the effectiveness of our simple image synthesis pipeline. 

\subsection{Results of Modality-aware Understanding}
From the results in \cref{tab: vvm-und}, we can observe further improvements on modality understanding tasks. 
These tasks focus on understanding the modality context and linking the modality knowledge and common sense, which is more associated with the language model in LMM. As shown in \cref{tab: vvm-und}, our VVM-tuning enhanced the ability to utilize modality context and reinterprete the perceptual features by a 3.8\% average improvement across all models and modalities. 
Similar to before, among all baselines, LLaVA-1.5 has the greatest improvement of 5.6\%, and the 4B Qwen-3-VL has the smallest improvement of 2.0\%. 
This result further confirms the possibility of generalization across diverse unseen visual modalities.
Tested LMMs demonstrate strong potential for learning from synthesized modalities and adapting to the other unseen visual modalities.
\begin{table}[tb]
\caption{The results of the Modality-aware Understanding subset of VVM-bench with modality contexts. “$\Delta$” represents accuracy difference between the model we tuned and the baselines. “Average” shows the average accuracy and difference across all tested modalities. “Average $\Delta$” shows the average improvements across all tested models for each visual modality. }
\label{tab: vvm-und}
\resizebox{\textwidth}{!}{\begin{tabular}{c|c|cccccc|c}
\toprule
\multirow{2}{*}{Model}& \multirow{2}{*}{Size} & \multicolumn{3}{c}{Collected Visual Modalities} & \multicolumn{3}{c|}{Fabricated Visual Modalities} 
& \multirow{2}{*}{Average}\\
& & Thermal & Depth & X-Ray
& Optical Flow & Synthesized 1 & Synthesized 2 & \\ \midrule
LLaVA-1.5 & 7b & 77.8\% & 80.3\% & 72.4\% & 76.7\% & 84.1\% & 84.8\% & 79.4\% \\
LLaVA-1.5~(Ours) & 7b & 82.5\% & 81.2\% & 79.3\% & 89.2\% & 87.6\% & 89.8\% & 84.9\% \\
$\Delta$ & - & \cdelta{4.7\%} & \cdelta{0.9\%} & \cdelta{6.9\%} & \cdelta{12.5\%} & \cdelta{3.5\%} & \cdelta{5.0\%} & \cdelta{5.6\%} \\
Qwen-2.5-VL & 3b & 83.8\% & 85.5\% & 83.5\% & 89.6\% & 86.9\% & 91.0\% & 86.7\% \\
Qwen-2.5-VL~(Ours) & 3b & 89.3\% & 89.5\% & 85.9\% & 93.8\% & 91.0\% & 93.8\% & 90.6\% \\
$\Delta$ & - & \cdelta{5.5\%} & \cdelta{4.0\%} & \cdelta{2.4\%} & \cdelta{4.2\%} & \cdelta{4.1\%} & \cdelta{2.8\%} & \cdelta{3.8\%} \\
Qwen-2.5-VL & 7b & 86.4\% & 83.4\% & 84.1\% & 84.6\% & 88.8\% & 89.8\% & 86.2\% \\
Qwen-2.5-VL~(Ours) & 7b & 88.5\% & 90.2\% & 85.9\% & 93.8\% & 91.9\% & 93.5\% & 90.6\% \\
$\Delta$ & - & \cdelta{2.1\%} & \cdelta{6.8\%} & \cdelta{1.8\%} & \cdelta{9.2\%} & \cdelta{3.1\%} & \cdelta{3.7\%} & \cdelta{4.5\%} \\
Qwen-3-VL & 4b & 87.3\% & 83.8\% & 84.3\% & 90.0\% & 90.0\% & 91.8\% & 87.9\% \\
Qwen-3-VL~(Ours) & 4b & 89.6\% & 87.0\% & 88.0\% & 92.1\% & 90.5\% & 92.0\% & 89.9\% \\
$\Delta$ & - & \cdelta{2.3\%} & \cdelta{3.2\%} & \cdelta{3.7\%} & \cdelta{2.1\%} & \cdelta{0.5\%} & \cdelta{0.2\%} & \cdelta{2.0\%} \\
Qwen-3-VL & 8b & 89.3\% & 87.2\% & 83.1\% & 90.8\% & 87.9\% & 91.5\% & 88.3\% \\
Qwen-3-VL~(Ours) & 8b & 91.6\% & 91.5\% & 89.8\% & 91.7\% & 90.3\% & 93.8\% & 91.5\% \\
$\Delta$ & - & \cdelta{2.3\%} & \cdelta{4.3\%} & \cdelta{6.7\%} & \cdelta{0.9\%} & \cdelta{2.4\%} & \cdelta{2.3\%} & \cdelta{3.2\%} \\ \midrule
Average $\Delta$ & - & \cdelta{3.4\%} & \cdelta{3.8\%} & \cdelta{4.3\%} & \cdelta{5.8\%} & \cdelta{2.7\%} & \cdelta{2.8\%} & \cdelta{3.8\%}\\
\bottomrule
\end{tabular}}
\vspace{-0.3 in}
\end{table}

Among the evaluated modalities, the Optical Flow fabricated by us is the most improved, and the other three real modalities are next to it; our synthesized modalities get the lowest improvements. 
This shows that a synthetic gap still exists between made-up physical meaning and realistic sensing from the physical world.
However, this gap does not reflect in the perception subset, indicating that modality meanings are more specific and harder to imitate than their visual appearances.
Nevertheless, the 2.7\%-5.8\% improvements still support that the model can learn from synthesized modality contexts to generalize to real visual modalities and keep the potential for completely unseen visual modalities.

\subsection{Ablation Study}

We also conduct ablation experiments to further observe the effectiveness of our training strategy. The first ablation demonstrates the effect of introducing modality context on real visual modalities. The results in \cref{tab: abl-ctx} indicate that the modality context is important for modality understanding when encountering unseen modalities, as removing modality context leads to a 32.75\% decrease at most. 
However, as the data scales of modern LMMs are continuously expanding, common non-RGB modalities like Thermal and X-ray are inevitably involved in their knowledge base. Thus, we observe a rising trend as the model size and date increase in Thermal and X-ray compared to Depth, even without modality context.
Also, the results confirm that our training does not inject modality knowledge during fine-tuning; instead, improved modality-unaware perception brought an increase in modality understanding even without modality context.
\begin{table}[tb]
\caption{The results of 3 real visual modalities in the understanding subset with/out modality contexts. “M.C.” is an abbreviation for Modality Context. “$\Delta$” represents the accuracy difference with or without modality context for each model and modality. “Average $\Delta$” demonstrates the average improvements of providing modality contexts for each modality across all models.}
\label{tab: abl-ctx} 
\resizebox{\textwidth}{!}{\begin{tabular}{c|c|ccccccccc}
\toprule
\multirow{2}{*}{Model}& \multirow{2}{*}{Size} & \multicolumn{3}{c}{Thermal} & 
\multicolumn{3}{c}{Depth} &
\multicolumn{3}{c}{X-Ray} \\
& & w/o M.C. & w/ M.C. & $\Delta$ 
& w/o M.C. & w/ M.C. & $\Delta$
& w/o M.C. & w/ M.C. & $\Delta$\\ \midrule
LLaVA-1.5 & 7b & 47.7\% & 77.8\% & \cdelta{30.2\%} & 40.7\% & 80.3\% & \cdelta{39.6\%} & 61.0\% & 72.4\% & \cdelta{11.5\%} \\
LLaVA-1.5~(Ours) & 7b & 63.1\% & 82.5\% & \cdelta{19.4\%} & 43.5\% & 81.2\% & \cdelta{37.6\%} & 73.0\% & 79.3\% & \cdelta{6.3\%} \\
Qwen-2.5-VL & 3b & 76.0\% & 83.8\% & \cdelta{7.8\%} & 54.4\% & 85.5\% & \cdelta{31.0\%} & 79.7\% & 83.5\% & \cdelta{3.8\%} \\
Qwen-2.5-VL~(Ours) & 3b & 79.9\% & 89.3\% & \cdelta{9.5\%} & 64.7\% & 89.5\% & \cdelta{24.8\%} & 84.4\% & 85.9\% & \cdelta{1.5\%} \\
Qwen-2.5-VL & 7b & 69.6\% & 86.4\% & \cdelta{16.8\%} & 46.3\% & 83.4\% & \cdelta{37.1\%} & 82.9\% & 84.1\% & \cdelta{1.2\%} \\
Qwen-2.5-VL~(Ours) & 7b & 74.5\% & 88.5\% & \cdelta{13.9\%} & 54.3\% & 90.2\% & \cdelta{35.9\%} & 84.1\% & 85.9\% & \cdelta{1.8\%} \\
Qwen-3-VL & 4b & 77.5\% & 87.3\% & \cdelta{9.8\%} & 47.9\% & 83.8\% & \cdelta{35.9\%} & 83.4\% & 84.3\% & \cdelta{0.9\%} \\
Qwen-3-VL~(Ours) & 4b & 73.8\% & 89.6\% & \cdelta{15.8\%} & 55.9\% & 87.0\% & \cdelta{31.1\%} & 86.9\% & 88.0\% & \cdelta{1.1\%} \\
Qwen-3-VL & 8b & 71.4\% & 89.3\% & \cdelta{17.9\%} & 53.8\% & 87.2\% & \cdelta{33.4\%} & 82.3\% & 83.1\% & \cdelta{0.8\%} \\
Qwen-3-VL~(Ours) & 8b & 79.1\% & 91.6\% & \cdelta{12.6\%} & 70.4\% & 91.5\% & \cdelta{21.1\%} & 88.2\% & 89.8\% & \cdelta{1.6\%} \\ \midrule
Average $\Delta$ & - & - & - & \cdelta{15.37\%} & - & - & \cdelta{32.75\%} & - & - & \cdelta{3.05\%} \\
\bottomrule
\end{tabular}}
\vspace{-0.15 in}
\end{table}

Further ablation provided a result about the effectiveness of two aspects of VVM-tuning. In \cref{tab: abl-1}, we conduct an ablation study on our tuned 7B Qwen-2.5-VL model to explore the influence of modality-unaware perception tasks and modality-aware understanding tasks. As shown in \cref{tab: abl-1}, both parts are necessary for complete modality understanding.
The modality-unaware perception tasks solely brought roughly half of the total improvement on both perception and understanding subsets.
And the modality-aware understanding training further enhances the improvements on the basis of modality-unaware perception.
However, without modality-unaware tasks, the perception base is too weak to learn the adaptability of modality understanding for unseen visual modalities. Therefore, the performance nearly drops back to baseline.

\begin{table}[tb]
\centering
\caption{Ablation results for the modality-unaware perception and modality-aware understanding in VVM-Tuning on Qwen 2.5-VL 7B model. “Perception” and “Understanding” represent modality-unaware perception tasks and modality-aware understanding tasks in VVM-Tuning, respectively. Instead, “P” and “U” represent the perception subset and understanding subset in VVM-Bench for evaluation. “Average” shows an average accuracy across all visual modalities. “{\scriptsize\cha}” and “{\scriptsize\gou}” indicate whether the corresponding task set is used or not used during training.}
\vspace{-0.02 in}
\label{tab: abl-1}
\resizebox{\textwidth}{!}{\begin{tabular}{cc|c|cccccc|c}
\toprule
\multirow{2}{*}{Perception} & \multirow{2}{*}{Understanding} & \multirow{2}{*}{Subset} & \multicolumn{3}{c}{Collected Visual Modalities} & \multicolumn{3}{c|}{Fabricated Visual Modalities}
& \multirow{2}{*}{Average} \\
& & & Thermal & Depth & X-Ray 
& Optical Flow & Synthesized 1 & Synthesized 2 & \\ \midrule
\multirow{2}{*}{\normalsize\cha} & \multirow{2}{*}{\normalsize\cha} & P & 74.5\% & 73.3\% & 75.7\% & 77.0\% & 87.1\% & 81.1\% & 78.1\% \\ 
 &  & U & 86.4\% & 83.4\% & 84.1\% & 84.6\% & 88.8\% & 89.8\% & 86.2\% \\ [5pt]
\multirow{2}{*}{\normalsize\gou} & \multirow{2}{*}{\normalsize\cha} & P & 75.3\% & 75.6\% & 79.6\% & 81.7\% & 90.7\% & 87.6\% & 81.7\% \\ 
 &  & U & 86.7\% & 82.5\% & 86.3\% & 92.9\% & 90.3\% & 91.3\% & 88.3\% \\ [5pt]
\multirow{2}{*}{\normalsize\cha} & \multirow{2}{*}{\normalsize\gou} & P & 76.8\% & 74.1\% & 75.4\% & 74.1\% & 85.8\% & 81.7\% & 78.0\% \\ 
 &  & U & 87.6\% & 85.4\% & 85.1\% & 92.1\% & 88.8\% & 90.3\% & 88.2\% \\ [5pt]
\multirow{2}{*}{\normalsize\gou} & \multirow{2}{*}{\normalsize\gou} & P & 79.8\% & 79.2\% & 79.6\% & 88.4\% & 91.2\% & 86.4\% & 84.1\% \\
 &  & U & 88.5\% & 90.2\% & 85.9\% & 93.8\% & 91.9\% & 93.5\% & 90.6\%\\
 \bottomrule
\end{tabular}}
\vspace{-0.15 in}
\end{table}
\section{Conclusion}

Through our exploration of unseen visual modality generalization of LMMs, we have demonstrated that the disentangled alignments of semantics and photographic perception are effective as a generalization foundation. Via fabricating images, the modality-unaware tasks equip the model with unified perception across multiple real and synthesized visual modalities. Furthermore, we introduce modality context in the prompt and enable the adaptation to unseen visual modalities through modality synthesis. Our experiments have shown that the training framework is effective on multiple base models and helps them improve 8.2\% and 3.8\% performance on perception and understanding respectively without in-modality data. Therefore, the results confirmed that generalizing among versatile unseen visual modalities is possible for LMMs with enough perception and modality contexts.
We also propose a benchmark to evaluate LMMs' general abilities on diverse visual modalities. We hope our efforts could facilitate the research on the visual modality adaptation of LMMs.

This attempt also gives us a new view of understanding the underlying mechanism of LMMs' vision. By distinguishing the differences and commonalities between visual modalities, we have the idea of separating shared semantics and photographic elements, which leads to a rethinking of machine vision: Do LMMs understand the difference between the image and the real world behind it? Or is its “real world” just the image?
We wish this could inspire others to a deeper understanding of LMMs' visual abilities and continue to explore the boundaries of LMMs' vision.
\section*{Acknowledgements}
This work was supported by the National Key R\&D Program of China under Grant No.~2022YFA1004100 and the National Natural Science Foundation of China (NSFC) under Grant No.~62501191.

Besides, we express our gratitude to the anonymous reviewers for their invaluable suggestions and comments.
We would like to thank all members of the ILL Lab for their helpful discussions and the computing resources supported by the ILL Lab, Faculty of Computing, HIT. Also, thanks to Sangyun Chung and other authors of VS-TDX for the accessibility of Non-RGB data.

%
%
\bibliographystyle{splncs04}
\bibliography{main}
\clearpage
\setcounter{page}{1}
\title{Supplementary Material}
\renewcommand{\thesection}{\Alph{section}}
\renewcommand{\thetable}{A\arabic{table}}
\renewcommand{\thefigure}{A\arabic{figure}}
\renewcommand{\thealgorithm}{A\arabic{algorithm}}

\def\tabularxcolumn#1{m{#1}}

\author{}

\authorrunning{Shihao Yuan, Yuanze Li,~\etal}

\institute{}

\maketitle

\section{Additional Results}

\subsection{Quantity Results on VS-TDX}
As discussed in the main paper, we have collected 3 modalities from VS-TDX~\cite{mspr} and modified the prompt formation to insert modality context. The results in \cref{tab: mspr} are evaluated under the original protocol of VS-TDX without our modification.
The ``Perception Mean'' and ``Understanding Mean'' are calculated by averaging the accuracies of perception tasks~(the first four tasks), and the understanding tasks~(the last two tasks), following VS-TDX.
Our modification has been shown in \cref{tab: a_mod}, where the blue text represents the modality contexts added by us, and the rest of the texts remain unchanged as VS-TDX.

\begin{table}[h]
\centering
\caption{The results under the evaluation protocol of VS-TDX.}
\label{tab: mspr}
\resizebox{\textwidth}{!}{\begin{tabular}{cccccc
>{\columncolor[HTML]{EFEFEF}}c cc
>{\columncolor[HTML]{EFEFEF}}c}
\toprule
Model & Modality & Exsistence & Counting & Position & Description & \begin{tabular}[c]{@{}c@{}}Perception \\ Mean\end{tabular} & \begin{tabular}[c]{@{}c@{}}Contextual \\ Understanding\end{tabular} & \begin{tabular}[c]{@{}c@{}}Sensor \\ Understanding\end{tabular} & \begin{tabular}[c]{@{}c@{}}Understanding \\ Mean\end{tabular}\\ \midrule
\multirow{3}{*}{LLaVA-1.5 7B} & Thermal & 57.1\% & 32.7\% & 68.8\% & 59.9\% & 54.6\% & 73.1\% & 38.4\% & 55.8\% \\ 
 & Depth & 69.7\% & 23.2\% & 55.2\% & 74.7\% & 55.7\% & 68.9\% & 22.6\% & 45.8\% \\
 & X-ray & 64.6\% & 35.3\% & 54.6\% & 73.4\% & 57.0\% & 76.2\% & 52.0\% & 64.1\% \\ \midrule
\multirow{3}{*}{LLaVA-1.5 7B~(Ours)} & Thermal & 71.5\% & 57.3\% & 75.0\% & 69.9\% & 68.4\% & 79.3\% & 58.5\% & 68.9\% \\
 & Depth & 74.5\% & 50.0\% & 67.3\% & 77.6\% & 67.3\% & 76.3\% & 20.6\% & 48.5\% \\
 & X-ray & 66.2\% & 60.2\% & 67.3\% & 78.7\% & 68.1\% & 77.9\% & 68.3\% & 73.1\% \\ \midrule
\multirow{3}{*}{Qwen-2.5-VL 3B} & Thermal & 75.7\% & 46.7\% & 72.7\% & 78.4\% & 68.4\% & 83.5\% & 73.9\% & 78.7\% \\
 & Depth & 75.5\% & 40.0\% & 62.9\% & 80.0\% & 64.6\% & 74.3\% & 40.6\% & 57.4\% \\
 & X-ray & 75.0\% & 34.1\% & 61.8\% & 82.4\% & 63.3\% & 82.3\% & 77.1\% & 79.7\% \\ \midrule
\multirow{3}{*}{Qwen-2.5-VL 3B~(Ours)} & Thermal & 83.5\% & 70.4\% & 78.1\% & 85.2\% & 79.3\% & 87.1\% & 77.8\% & 82.4\% \\
 & Depth & 80.3\% & 59.5\% & 75.7\% & 83.8\% & 74.8\% & 80.2\% & 53.9\% & 67.0\% \\
 & X-ray & 76.6\% & 72.3\% & 76.9\% & 87.3\% & 78.3\% & 85.7\% & 83.2\% & 84.5\% \\ \midrule
\multirow{3}{*}{Qwen-2.5-VL 7B} & Thermal & 78.9\% & 47.2\% & 77.3\% & 80.2\% & 70.9\% & 81.6\% & 66.2\% & 73.9\% \\
 & Depth & 78.2\% & 51.6\% & 70.5\% & 82.1\% & 70.6\% & 71.5\% & 28.7\% & 50.1\% \\
 & X-ray & 78.9\% & 44.6\% & 78.1\% & 92.6\% & 73.5\% & 84.9\% & 80.9\% & 82.9\% \\ \midrule
\multirow{3}{*}{Qwen-2.5-VL 7B~(Ours)} & Thermal & 79.7\% & 68.3\% & 86.7\% & 82.7\% & 79.4\% & 86.7\% & 71.0\% & 78.9\% \\
 & Depth & 83.2\% & 60.5\% & 78.4\% & 84.2\% & 76.6\% & 80.2\% & 36.2\% & 58.2\% \\
 & X-ray & 79.0\% & 71.1\% & 79.3\% & 90.6\% & 80.0\% & 85.3\% & 83.0\% & 84.2\% \\ \midrule
\multirow{3}{*}{Qwen-3-VL 4B} & Thermal & 86.4\% & 60.3\% & 75.0\% & 84.2\% & 76.5\% & 84.1\% & 75.6\% & 79.9\% \\
 & Depth & 88.2\% & 51.6\% & 72.3\% & 86.9\% & 74.7\% & 79.9\% & 25.5\% & 52.7\% \\
 & X-ray & 78.5\% & 48.6\% & 82.1\% & 93.0\% & 75.5\% & 85.7\% & 81.2\% & 83.5\% \\ \midrule
\multirow{3}{*}{Qwen-3-VL 4B~(Ours)} & Thermal & 85.1\% & 75.4\% & 84.4\% & 82.5\% & 81.8\% & 90.0\% & 69.2\% & 79.6\% \\
 & Depth & 89.2\% & 60.0\% & 82.6\% & 92.6\% & 81.1\% & 87.0\% & 34.1\% & 60.5\% \\
 & X-ray & 80.0\% & 63.1\% & 81.7\% & 92.2\% & 79.2\% & 88.2\% & 85.7\% & 86.9\% \\ \midrule
\multirow{3}{*}{Qwen-3-VL 8B} & Thermal & 85.4\% & 60.8\% & 78.1\% & 81.9\% & 76.6\% & 87.1\% & 66.9\% & 77.0\% \\
 & Depth & 86.1\% & 53.2\% & 74.7\% & 87.4\% & 75.3\% & 78.0\% & 36.9\% & 57.4\% \\
 & X-ray & 76.4\% & 42.2\% & 76.5\% & 93.4\% & 72.1\% & 83.8\% & 80.9\% & 82.3\% \\ \midrule
\multirow{3}{*}{Qwen-3-VL 8B~(Ours)} & Thermal & 83.9\% & 73.9\% & 77.3\% & 84.8\% & 80.0\% & 92.2\% & 75.3\% & 83.8\% \\
 & Depth & 88.4\% & 68.4\% & 82.0\% & 90.0\% & 82.2\% & 87.0\% & 58.8\% & 72.9\% \\
 & X-ray & 79.4\% & 69.5\% & 81.3\% & 93.9\% & 81.0\% & 88.1\% & 88.3\% & 88.2\% \\ 
 \bottomrule
\end{tabular}}
\end{table}


\begin{table}[t]
    \centering
    \caption{An example of our modification, we added the \textcolor[HTML]{5B9BD5}{modality context} to the original prompt of VS-TDX for understanding tasks.}
    \label{tab: a_mod}
    \vspace{-0.1in}
    \begin{tcolorbox}[colback=gray!5!white,colframe=white]
        \begin{minipage}[t]{\textwidth}
        \begin{tcolorbox}[title=\textbf{Question}]
            \textcolor[HTML]{5B9BD5}{Thermal images visualize infrared radiation emitted by objects using heat-sensing sensors. They can be used to analyze temperature distribution, detect objects, and inspect equipment conditions. Usually, the color of thermal images represents for temperatures, for example, high temperature usually appears reddish or bright in thermal image. 
            Here shows a thermal image, please answer question according to above information.}\\ 
            <image>\\
            Question: What is the temperature characteristic of the figure in the center of the image? Choices: \\
            A. The figure has the same temperature as the surroundings. \\
            B. The figure is warmer than the surroundings. \\
            C. The figure is cooler than the surroundings. \\
            D. The figure is not distinguishable temperature-wise.\\
            Please directly answer the question and provide the correct option letter, e.g., A, B, C, D.
        \end{tcolorbox}
        \end{minipage}
        \begin{tcolorbox}[title=\textbf{Answer},]
            B
        \end{tcolorbox}
    \end{tcolorbox}
\end{table}
\subsection{Data Scaling on Qwen-2.5-VL 7B}
Results in \cref{tab: data_scaling_p} reveal an ascending trend with increasing data scales. However, scaling the number of samples is less meaningful than increasing image styles and task diversity for generalization, which we expect in future work.
\begin{table}[htb]
\vspace{-0.1 in}
\centering
\setlength{\tabcolsep}{4pt}
\caption{Perception results for data scales on Qwen-2.5-VL 7B. }
\label{tab: data_scaling_p}
\resizebox{\linewidth}{!}{
\begin{tabular}{c|cccccc|c}
\toprule
Data Scale & Thermal & Depth & X-Ray & Optical Flow & Synthesized 1 & Synthesized 2 & Mean \\ \midrule
baseline & 74.5\% & 73.3\% & 75.7\% & 77.0\% & 87.1\% & 81.1\% & 78.1\% \\
25\% & 78.3\% & 77.3\% & 78.7\% & 83.3\% & 89.7\% & 84.1\% & 81.9\% \\
50\% & 76.7\% & 78.3\% & 78.4\% & 77.8\% & 88.8\% & 85.0\% & 80.8\% \\
75\% & 77.7\% & 79.1\% & 81.5\% & 82.8\% & 91.3\% & 86.1\% & 83.1\% \\
100\% & 79.8\% & 79.2\% & 79.6\% & 88.4\% & 91.2\% & 86.4\% & 84.1\% \\
\bottomrule
\end{tabular}}
\vspace{-0.1 in}
\end{table}

\subsection{Target-tuning with a Samll amount of Real Data}
We provide the results of tuning Qwen-2.5-VL 7B with TDX data~(Extra \textbf{T}hermal \textbf{D}epth, and \textbf{X}-Ray data, which is not in VVM-Bench) in \cref{tab: target_tuning}. With 600 samples~(each 200 for Thermal, Depth, and X-Ray) and 10 epochs, the performance is close to our method. With 3600 samples~(1200 per modality) and 10 epochs, the performance on these modalities surpasses ours. However, the performance on the other modalities remains close to baseline, indicating that simply training with several modalities benefits few in generalization. 
That makes our work meaningful, as we don't expect that synthetic modality tuning could replace real modality tuning. Instead, our method aims to enhance LMMs when real modality tuning is unavailable.
\begin{table}[htb]
\vspace{-0.1 in}
\centering
\setlength{\tabcolsep}{2pt}
\caption{Perception results with small amount of TDX data on Qwen-2.5-VL 7B and Qwen-3-VL 8B.}
\label{tab: target_tuning}
\resizebox{\linewidth}{!}{
\begin{tabular}{c|cccccc|c}
\toprule
Model & Thermal & Depth & X-Ray & Optical Flow & Synthesized 1 & Synthesized 2 & Mean \\ \midrule
Qwen-2.5-VL & 74.5\% & 73.3\% & 75.7\% & 77.0\% & 87.1\% & 81.1\% & 78.1\% \\
Qwen-2.5-VL~(Ours) & 79.8\% & 79.2\% & 79.6\% & 88.4\% & 91.2\% & 86.4\% & 84.1\% \\
Qwen-2.5-VL~(600) & 79.5\% & 78.2\% & 80.0\% & 75.4\% & 85.2\% & 79.6\% & 79.6\% \\
$\Delta$~(-ours) & \ddelta{-0.3\%} & \ddelta{-1.0\%} & \cdelta{0.4\%} & \ddelta{-13.0\%} & \ddelta{-6.0\%} & \ddelta{-6.8\%} & \ddelta{-4.5\%} \\
Qwen-2.5-VL~(3600) & 87.4\% & 82.2\% & 85.2\% & 76.5\% & 86.4\% & 82.6\% & 83.4\% \\
$\Delta$~(-ours) & \cdelta{7.6\%} & \cdelta{3.0\%} & \cdelta{5.6\%} & \ddelta{-11.9\%} & \ddelta{-4.8\%} & \ddelta{-3.8\%} & \ddelta{-0.7\%} \\ 
$\Delta$~(-baseline) & \cdelta{12.9\%} & \cdelta{8.9\%} & \cdelta{9.5\%} & \ddelta{-0.5\%} & \ddelta{-0.7\%} & \cdelta{1.5\%} & \cdelta{5.3\%} \\ 
\bottomrule
\end{tabular}}
\vspace{-0.3 in}
\end{table}

\subsection{Quality Results}
We observe the difference in the behavior of the baselines and our tuned model and display some examples here. Quality result 1 in \cref{tab: q_res_1}, serving as a failure case, demonstrates an unnecessary preference for \textcolor{orange}{guessing the visual modality} of the non-RGB images in the base model, which is unfortunately inherited by our tuned model. This preference comes from the insufficient quantity and diversity of non-RGB modalities during the visual-language alignments, which may require complete realignment with diverse non-RGB images. However, our tuned model still exhibits a more detailed \textcolor{blue}{perception of colors}.

Quality result 2 in \cref{tab: q_res_2} shows the ability to connect the modality context with visual perception. The base model \textcolor{red}{wrongly assumes} the red area in the center of the plant represents high fungal signals, indicating an inability to link visual perceptions with the physical meanings in the modality context. Meanwhile, our tuned model \textcolor{ForestGreen}{correctly links} the color of areas to the modality context, resulting in a more consistent understanding.

\begin{table}[H]
    \centering
    \caption{Quality result 1: modality-unaware description task on Qwen-2.5 VL 3B and our tuned model.}
    \label{tab: q_res_1}
    \vspace{-0.1in}
    \begin{tcolorbox}[
        title=\textcolor{white}{\textbf{Quality Result 1}}
    ]
    \begin{minipage}[t]{\textwidth}
        \centering
        \includegraphics[width=0.9\textwidth]{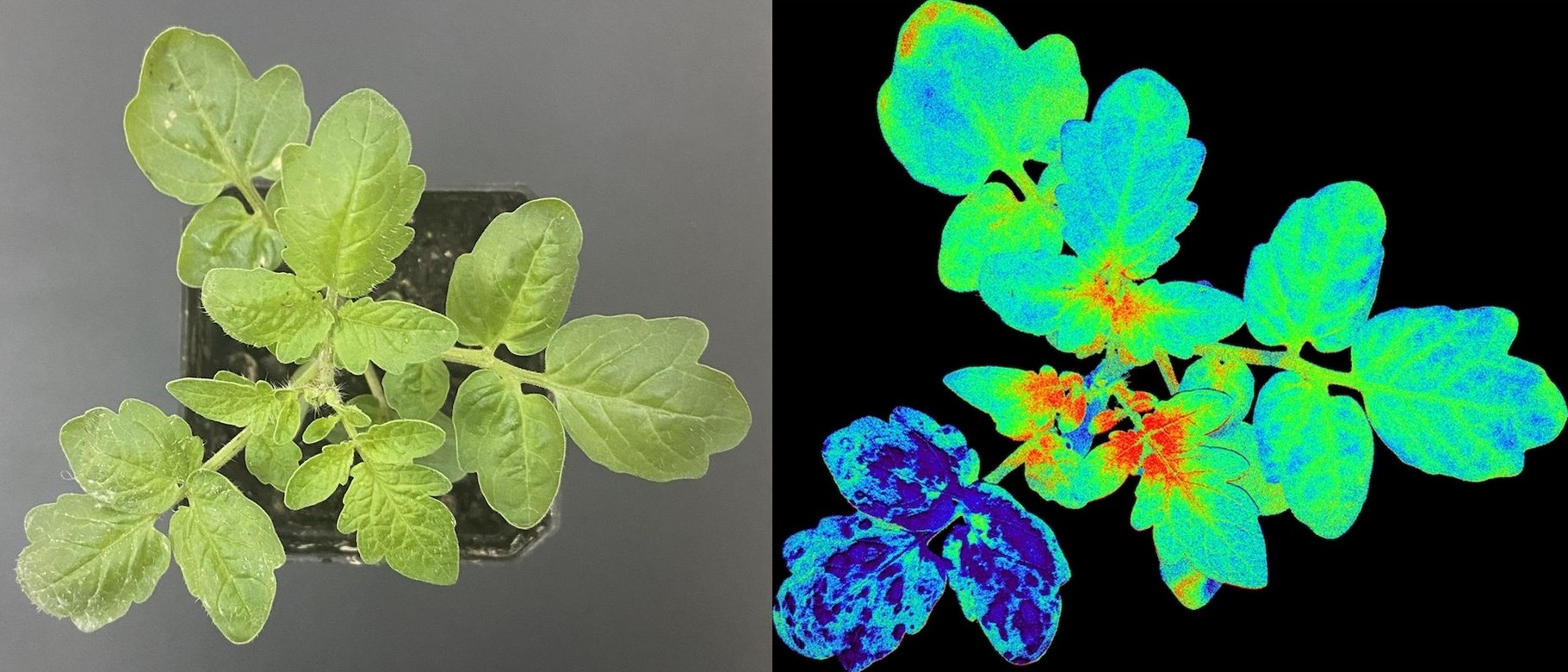}
    \end{minipage}\vspace{0.1in}
    
    \begin{minipage}[t]{\textwidth}
        \begin{tcolorbox}[title={\textbf{Instruction}}]
        \selectfont\ttfamily
        Describe the right image.
        \end{tcolorbox}
    \end{minipage}

    \begin{tcolorbox}[title=\textbf{Model Response}]
    \textbf{Qwen-2.5 VL 3B:}\\
    \selectfont\ttfamily
    The right image is a \textcolor{ForestGreen}{color-coded representation} of a plant, likely an \textcolor{orange}{infrared or thermal imaging scan. The colors indicate different temperature levels, with warmer areas appearing in red and cooler areas in blue.} The plant appears to have \textcolor{orange}{a healthy appearance with no visible signs of disease or damage}. The detailed coloration helps in identifying specific regions of the plant that might require further attention or monitoring.
    \tcblower
    \textbf{Qwen-2.5 VL 3B~(Ours):}\\
    \selectfont\ttfamily
    The image on the right is a \textcolor{orange}{thermal scan of a plant, showing heat signatures} in various colors such as \textcolor{blue}{blue, green, orange, and red}.
    \end{tcolorbox}
    \end{tcolorbox}
\end{table}

\begin{table}[H]
    \centering
    \caption{Quality result 2: modality-aware understanding on Qwen-2.5 VL 3B and our tuned model.}
    \label{tab: q_res_2}
    \vspace{-0.1in}
    \begin{tcolorbox}[
        title=\textcolor{white}{\textbf{Quality Result 2}}
    ]
    \begin{minipage}[t]{\textwidth}
        \centering
        \includegraphics[width=0.9\textwidth]{figures/014.jpg}
    \end{minipage}\vspace{0.1in}
    
    \begin{minipage}[t]{\textwidth}
        \begin{tcolorbox}[title={\textbf{Instruction}}]
        \selectfont\ttfamily
        Context:\\
        The image is visualizing the fungal distribution. The blue color means more fungal signals detected, and the red area means less fungal signals.\\
        Question:\\
        What can we learn from the right image?
        \end{tcolorbox}
    \end{minipage}

    \begin{tcolorbox}[title=\textbf{Model Response}]
    \textbf{Qwen-2.5 VL 3B:}\\
    \selectfont\ttfamily
    We can see that there is a high concentration of fungal signals \textcolor{red}{in the center of the plant}.
    \tcblower
    \textbf{Qwen-2.5 VL 3B~(Ours):}\\
    \selectfont\ttfamily
    The right image shows that the plant has areas with \textcolor{ForestGreen}{high fungal activity (blue) and low fungal activity (red)}. This suggests that the plant may have regions where it is more susceptible to fungal infection.
    \end{tcolorbox}
    \end{tcolorbox}
    \vspace{-0.3in}
\end{table}

The question in \cref{tab: q_res_3} tests the ability of modality-unaware perception by asking about the existence of objects in the image. The base model, Qwen-2.5 VL 7B, is still trying to \textcolor{orange}{guess the modality} of the image and \textcolor{red}{refuses to answer} the question by requiring more contexts. However, it is enough to determine the existence of the forest just by viewing the image itself, which is \textcolor{ForestGreen}{successfully achieved} by our tuned model.

The \cref{tab: q_res_4} shows an open understanding task with modality context. The base model answers with a lengthy statement that falsely believes the trees are \textcolor{red}{concentrated} and \textcolor{red}{located in a lower altitude} because of their \textcolor{red}{green or blue colors}. This error clearly reflects how erroneous perception affects the understanding, even with modality contexts, while our tuned model exhibits a correct understanding based on robust perception instead.
\begin{table}[H]
    \vspace{-0.2in}
    \centering
    \caption{Quality result 3: modality-unaware perception on Qwen-2.5 VL 7B and our tuned model.}
    \label{tab: q_res_3}
    \vspace{-0.1in}
    \begin{tcolorbox}[
        title=\textcolor{white}{\textbf{Quality Result 3}}
    ]
    \begin{minipage}[t]{\textwidth}
        \centering
        \includegraphics[width=0.7\textwidth]{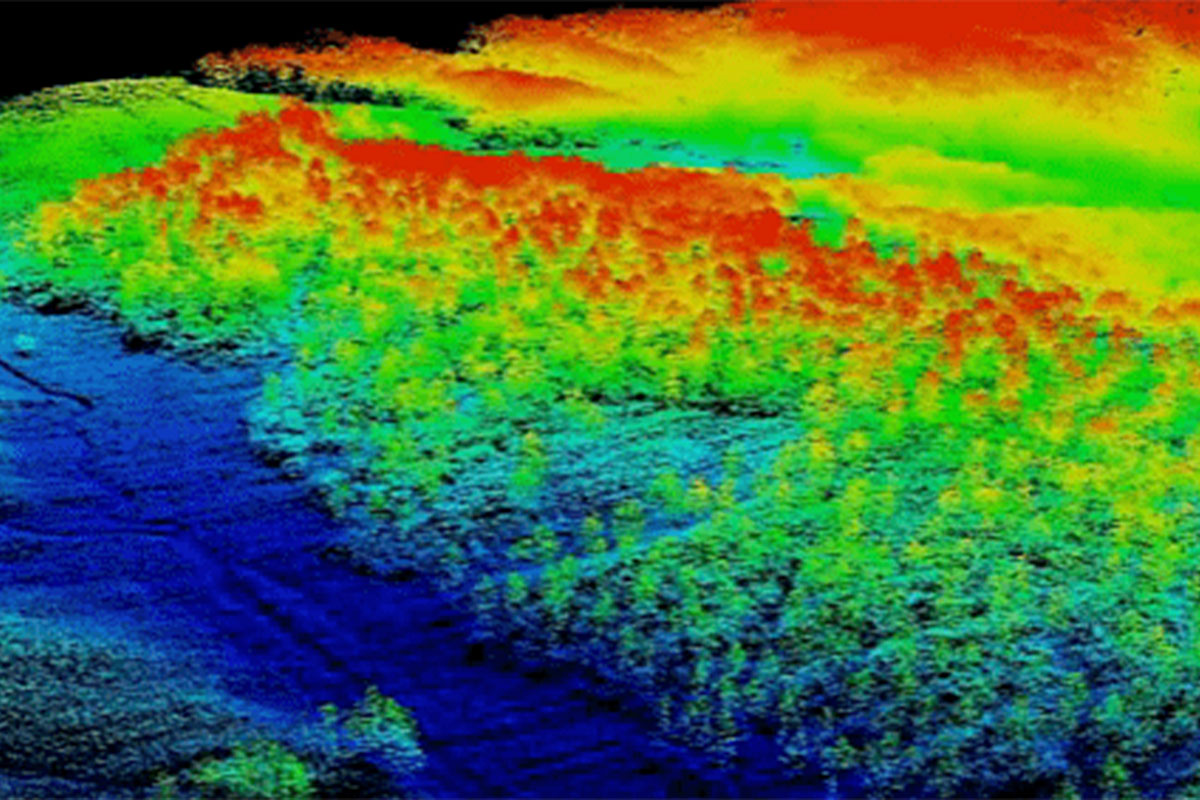}
    \end{minipage}\vspace{0.1in}
    
    \begin{minipage}[t]{\textwidth}
        \begin{tcolorbox}[title={\textbf{Instruction}}]
        \selectfont\ttfamily
        Are there any forest in the image?
        \end{tcolorbox}
    \end{minipage}
    \begin{tcolorbox}[title=\textbf{Model Response}]
    \textbf{Qwen-2.5 VL 7B:}\\
    \selectfont\ttfamily
    The image appears to be a topographical map or a digital elevation model (DEM) of an area, \textcolor{red}{not a photograph of a forest}. The colors represent different elevations and \textcolor{orange}{possibly vegetation types}, but they do not depict a forest in the traditional sense. The blue areas likely represent lower elevations, while the green and yellow areas could indicate higher elevations and potentially areas with vegetation. \textcolor{red}{However, without additional context or a legend, it's not possible to definitively state that there is a forest present.} If you have more information about the location or the purpose of this image, it might help in identifying specific features like forests.
    \tcblower
    \textbf{Qwen-2.5 VL 7B~(Ours):}\\
    \selectfont\ttfamily
    \textcolor{ForestGreen}{Yes, there is a forest visible in the image.}
    \end{tcolorbox}
    \end{tcolorbox}
\end{table}
\begin{table}[H]
    \centering
    \caption{Quality result 4: modality-aware understanding on Qwen-2.5 VL 7B and our tuned model.}
    \label{tab: q_res_4}
    \vspace{-0.1in}
    \begin{tcolorbox}[
        title=\textcolor{white}{\textbf{Quality Result 4}}
    ]
    \begin{minipage}[t]{\textwidth}
        \centering
        \includegraphics[width=0.7\textwidth]{figures/020.jpg}
    \end{minipage}\vspace{0.1in}
    
    \begin{minipage}[t]{\textwidth}
        \begin{tcolorbox}[title={\textbf{Instruction}}]
        \selectfont\ttfamily
        Context:\\
        The image shows an altitude map of a landscape. The color closer to red means higher altitude. The color closer to blue means lower altitude.\\
        Question:\\
        What can we learn about the trees in the image?
        \end{tcolorbox}
    \end{minipage}
    \begin{tcolorbox}[title=\textbf{Model Response}]
    \textbf{Qwen-2.5 VL 7B:}\\
    \selectfont\ttfamily
    From the image, which appears to be an altitude map of a landscape, we can infer some characteristics about the trees based on their distribution and the colors representing different altitudes:\\
    Tree Distribution: The trees seem to \textcolor{red}{be concentrated in certain areas rather than being uniformly spread across the landscape}. This suggests that there might be specific environmental conditions (such as soil type, water availability, or sunlight) that favor tree growth in those areas.\\
    Altitude Influence: The trees appear to be more prevalent in areas with \textcolor{red}{lower altitudes, as indicated by the blue and green colors}. This could imply that the trees are adapted to growing in these lower-altitude regions, possibly due to factors such as temperature, precipitation, or soil composition.\\
    \end{tcolorbox}
    \end{tcolorbox}
\end{table}
\begin{table}[H]
    \begin{tcolorbox}
    \begin{tcolorbox}
    \selectfont\ttfamily
    Topography: The presence of trees in specific areas might also indicate that these areas have more favorable topographical features for tree growth, such as gentle slopes or areas where water collects.\\
    Vegetation Patterns: The distinct patches of different colors suggest that there might be variations in vegetation types or densities. The transition from one color to another could indicate changes in the type of vegetation or the health of the trees.\\
    Overall, the image provides insights into the relationship between tree distribution and altitude, suggesting that trees are influenced by both topographical and environmental factors in this landscape.
    \tcblower
    \textbf{Qwen-2.5 VL 7B~(Ours):}\\
    \selectfont\ttfamily
    The trees are located at \textcolor{ForestGreen}{a medium altitude}, as indicated by the \textcolor{blue}{orange and yellow colors} surrounding them.
    \end{tcolorbox}
    \end{tcolorbox}
\end{table}

\subsection{Discussion of Limitations}
As discussed before, the generalization to unseen visual modalities is based on a modality-unaware perception, from which the visual-language alignment is inherited from RGB vision. 
Our tuning only disentangles that alignment by semantics and photographics, not executing a completely new alignment or injecting new knowledge. 
Besides, our modality context only introduces the mapping of physical meanings, without injecting new knowledge or abilities into the model.
That means the generalization to unseen modalities is limited within the existing knowledge and abilities of known objects learned by base LMMs.
To be honest, that would constrain the application of such generalization, as the non-RGB images are usually highly related to specific downstream tasks, which a general LMM could not solve without specialized domain knowledge. 
Although how such new knowledge could be absorbed by the model from the context during inference remains an unexplored challenge, the current research on LMMs with long-CoT and reinforcement learning still demonstrates a promising prospect. 

Meanwhile, the evaluation format of VQA we employed in VVM-Bench is quite simple and is unable to measure the complex reasoning ability of the tested LMMs. While it's enough to serve as a perception-evaluating metric, it's easy to be hacked by guessing the answer from options in complex understanding tasks, especially with competent contexts.
We are looking forward to further development on multi-hop inference and complex context reasoning, as well as benchmarks and evaluation metrics in the direction of generalization to unseen visual modalities.
We believe our method could serve as a perceptual basis for further exploration.

\section{Evaluation Details}

\subsection{Evaluation Templates}
The templates of prompts used in evaluation are shown in \cref{tab: eval_template} in our VVM-Bench. Basically, the prompts are composed of four parts: image, question, options, and a format instruction for calculating accuracy. For understanding tasks, modality contexts are inserted at ``<context>'' as shown in \cref{tab: eval_template}.
\begin{table}[H]
    \centering
    \caption{Prompt Templates of VVM-Bench in evaluation for all modalities and task sets.}
    \label{tab: eval_template}
    \begin{tabularx}{\textwidth}{ccX}
    \toprule
    Modality & Task set & \multicolumn{1}{c}{Template} \\ 
    \midrule
    \vspace{1em}
    \multirow{8}{*}{Collected Modalities} &
    Perception&
    <image>\newline
    <question>\newline
    <options>\newline
    Please directly answer the question and provide the correct option letter, e.g., A, B, C, D. \\
    \vspace{1em}
    & Understanding&
    <context>\newline
    <image>\newline
    <question>\newline
    <options>\newline
    Please directly answer the question and provide the correct option letter, e.g., A, B, C, D. \\
    \vspace{1em}
    \multirow{8}{*}{Fabricated Modalities} &
    Perception &
    <image>\newline
    <question>\newline
    <options>\newline
    Answer with the option's letter from the given choices directly. \\
    & Understanding&
    <image>\newline
    <context>\newline
    <question>\newline
    <options>\newline
    Answer with the option's letter from the given choices directly. \\
    \bottomrule
    \end{tabularx}
\end{table}
\subsection{Benchmark Details}
Our VVM-Bench contains 13317 VQA samples in total, including 6991 samples in the perception subset and 6326 samples in the understanding subset. The detailed modalities and tasks distribution is listed in \cref{tab: vvm_bench_stats}. All samples follow the templates shown in \cref{tab: eval_template}. 

\begin{table}[htb]
\centering
\setlength{\tabcolsep}{2pt}
\caption{Number of samples in VVM-Bench.}
\label{tab: vvm_bench_stats}
\resizebox{\linewidth}{!}{
\begin{tabular}{c|ccccccc}
\toprule
Modality & Thermal & Depth & X-Ray & Optical Flow & Synthesized 1 & Synthesized 2 & Total \\ 
\midrule
P & 1330 & 1873 & 1692 & 378 & 1018 & 700 & 6991 \\
U & 1386 & 2304 & 1575 & 240 & 421 & 400 & 6326 \\
Total & 2716 & 4177 & 3267 & 618 & 1439 & 1100 & 13317 \\ 
\midrule
P & Cnt.(885) & Log.Inf.(1758) & Exis.(2235) & R.Pos.(1372) & Color(247) & Shape(247) & R.Size(247) \\
U & \multicolumn{2}{c}{Modal.Und.(3502)} & \multicolumn{2}{c}{Scene.Und.(2284)} & \multicolumn{2}{c}{Event Ext.(261)} & Modal.Reas.(279) \\
\bottomrule
\end{tabular}}
\end{table}

\begin{table}[H]
    \centering
    \caption{Modality Contexts of VVM-Bench in evaluation. ``Synthesized Context'' means the context is fabricated by us. }
    \label{tab: eval_mc}
    \begin{tabularx}{\textwidth}{cX}
    \toprule
    Modality & \multicolumn{1}{c}{Context} \\ 
    \midrule
    \vspace{1em}
    Thermal & 
    Thermal images visualize infrared radiation emitted by objects using heat-sensing sensors. They can be used to analyze temperature distribution, detect objects, and inspect equipment conditions. Usually, the color of thermal images represents for temperatures, for example, high temperature usually appears reddish or bright in thermal image.
    Here shows a thermal image, please answer question according to above information. \\
    \vspace{1em}
    Depth &
    Depth images visualize the distance between a sensor and objects in a scene by capturing depth information. They can be used to measure object dimensions, map environments in 3D, and assist in object recognition and navigation tasks. The color of depth image usually represents the depth or the distance between that point and the depth camera.Here shows a depth image, please answer question according to above information. \\
    \vspace{1em}
    Xray &
    X-ray images visualize the internal structures of objects by capturing the varying absorption of X-rays. They can be used to inspect internal components, identify structural defects, and analyze materials or biological tissues for diagnostic purposes. X-ray images usually shows different structures overlapping as the X-ray can go through the object, the more absorption of X-ray, more brighter in the image. Here shows a xray image, please answer question according to above information. \\
    \vspace{1em}
    Optical Flow &
    The image is an optical flow visualization. The color of the pixels represents for the movement of the object. The saturation of the color means the magnitude of the movement vector. The brighter the color, the larger the movement is. The paler color means smaller movement. The direction of the movement is represented by the hue of the color. Red means moving right, orange means moving lower right, yellow means moving down, green means moving lower left, blue means moving upper right, purple means moving up, pink means moving upper right. \\
    \vspace{1em}
    Synthesized 1 &
    Synthesized Context~(Hardness)\newline
    Synthesized Context~(Humidity) \\
    Synthesized 2 &
    Synthesized Context~(Hardness)\newline
    Synthesized Context~(Humidity)\newline
    Synthesized Context~(Heat Signals)\newline
    Synthesized Context~(Radiation)\newline
    Synthesized Context~(Reflective Level)\newline
    Synthesized Context~(Illuminance Level) \\
    \bottomrule
    \end{tabularx}
\end{table}

The detailed modality contexts are shown in \cref{tab: eval_mc}. Among the 6 modalities, 4 of them use modality contexts with real physical meanings. For the 2 synthesized modalities, we manufacture multiple meanings to compose 6 synthesized contexts as shown in \cref{tab: eval_syn_mc}. The synthesized modalities have multiple synthesized contexts among their samples, with the inclusion relationships listed in \cref{tab: eval_mc}.

\begin{table}[H]
    \centering
    \caption{Synthesized Modality Contexts of VVM-Bench for synthesized modalities.}
    \label{tab: eval_syn_mc}
    \begin{tabularx}{\textwidth}{cX}
    \toprule
    Modality & \multicolumn{1}{c}{Context} \\ 
    \midrule
    \vspace{1em}
    Hardness &
    The image visualizes the hardness of objects. Different colors represent different hardnesses. Red means solid hard, yellow means medium hardness, green means slightly soft, blue means medium soft, purple means very soft.\\
    \vspace{1em}
    Humidity & 
    The image visualizes the humidity in the air. Different colors represent different humidity. Red means low humidity, yellow means slightly dry, green means medium humidity, blue means humid, purple means high humidity. \\
    \vspace{1em}
    Heat Signals &
    The image visualizes the heat signals in the air. The colors in the picture range from black to white. The darker area means a weak signal, and the brighter area means a strong signal. \\
    \vspace{1em}
    Radiation &
    The image visualizes the radiation signals in the air. The colors in the picture range from red to yellow. The redder area means lower value, and the yellower area means higher value. \\
    \vspace{1em}
    Reflective Level &
    The image visualizes the reflective level of the object. The colors in the picture range from blue to red. The bluer area means low reflective level, and the redder area means high reflective level.\\
    \vspace{1em}
    Illuminance Llevel &
    The image visualizes the illuminance level in the area. The colors in the picture range from dark pink to white. The darker area means lower value, and the brighter area means higher value. \\
    \bottomrule
    \end{tabularx}
\end{table}

\section{Modality Synthesis Details}

\subsection{Image Synthesis}
The histogram adjusting method in \cref{fig: syn_pipe} is shown in \cref{alg: spline}.
We first generate 1-4 keypoints randomly in the value plane, ranging from 0 to 255. Then, we interpolate the cubic spline to form a map function $S(x)$ on $[0,255] \rightarrow [0,255]$. That function could change the histogram by changing the pixel value. However, random splines may overflow from $[0,255]$. Thus, we use rejection sampling to make sure the value range of $S(x)$ is valid.

\begin{figure}[ht]
    \centering
    \includegraphics[trim=3cm 1cm 3cm 1cm, clip=true, width=0.95\textwidth]{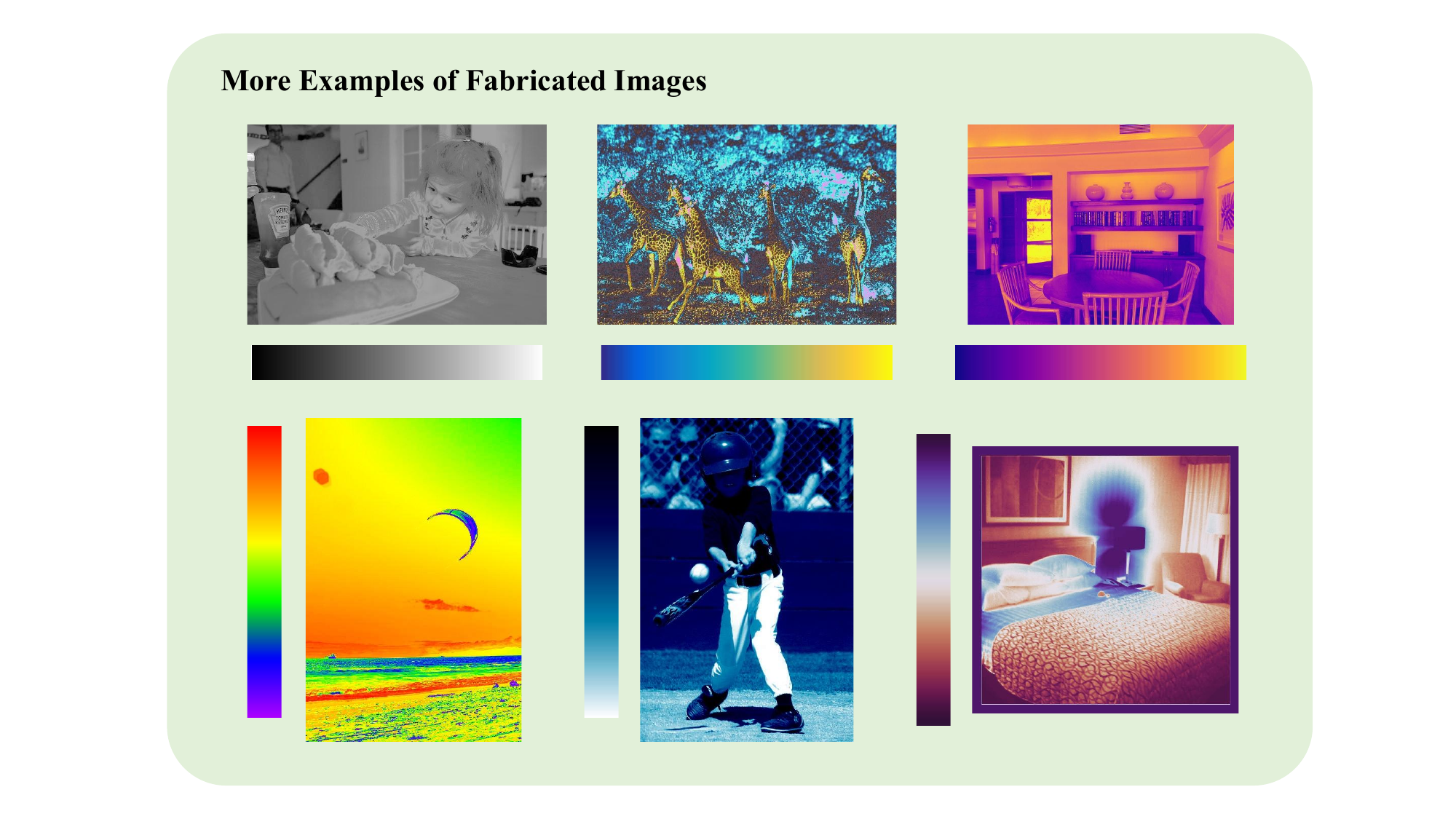}
    \caption{More examples of fabricated images produced by our image synthesis pipeline with different color maps.}
    \label{fig: syn_examples}
\end{figure}

The \cref{fig: syn_examples} shows more examples with corresponding colormap bars, indicating that the random splines are simple but effective in imitating the characteristics of non-RGB images. 


\begin{algorithm}[H]
  \caption{Adjust Histogram with Random Splines}
  \textbf{Input} Grayscale Image $I$; 
  \textbf{Hyperparameters} The max number of keypoints $N_{max}$; Max try times $T_{fail}$
  \begin{algorithmic}[1]
    \State Randomize the number of keypoints $N=Random(1,N_{max})$
    \For{cnt = 1, \dots, $T_{fail}$} 
    \State Generate $N$ random keypoints $(x_1,y_1)\dots(x_N,y_N)\in[0,255]^2$.
    \State Interpolate a cubic spline $S(x), x\in[0,255]$ using keypoints $(x_1,y_1)\dots(x_N,y_N)$.
    \If{$S(x)\in[0,255]$}
        \State \textbf{break}
    \Else
        \State The value range of $S(x)$ is illegal, reject $S(x)$.
    \EndIf
    \EndFor 
    \If{$\text{cnt}==T_{fail}$}
        \State $S(x)=x$
    \EndIf
    \State Adjust the histogram by mapping pixel value using the spline, $\hat{I}=S(I)$.
  \end{algorithmic}
  \textbf{Output} Adjusted Image $\hat{I}$
  \label{alg: spline}
\end{algorithm}


\subsection{Modality Context Template}

\begin{table}[H]
    \vspace{-0.1 in}
    \centering
    \caption{An example of templates used to fabricate modality context.}
    \label{tab: mc_template}
    \vspace{-0.1in}
    \begin{tcolorbox}[title={\textbf{Fabricated Modality Context Template}}]
    \selectfont\ttfamily
    The image is a special picture that records the <attr> of objects by some new imaging technique. Specificly the color of the image means different values of <attr>. <color\_code\_text> <range\_value>\\
    The meaning of colors can be described as below:\\
    <info\_color>.\\ \\
    <attr>: gravity level \\
    <color\_code\_text>: The colors in the picture range from <color\_left> to <color\_right>, follows a rainbow. The color closer to <color\_left> means lower value, and the color closer to <color\_right> means higher value. \\
    <range\_value>: The <attr> of the image could be divided into following levels: 0.2g, 0.5g, 1g...\\
    <info\_color>: The gravity level of <value> is represented by <color> color...\\
    \tcblower
    \selectfont\ttfamily
    An Example:\\
    Here shows an image captured by a special sensor. \\
    The image is a special picture that record the pressure level of objects by some new imaging technique. Specificly the color of the image means different value of pressure level. The colors in the picture range from red to purple, follows a rainbow. The color closer to red means lower value, and the color closer to purple means higher value. \\
    The meaning of colors can be described as below:\\
    The purple colored object in image has very high to very high pressure level. The color of object having high pressure level in image is blue color. The cyan color repesents for high to high in pressure level. The green colored object in image has medium to medium pressure level. The yellow colored object in image has low pressure level. The orange color repesents for low to low in pressure level. The red color is valued with very low to very low pressure level.
    \end{tcolorbox}
\end{table}

For the modality context synthesis, we employ both structured text and LMM-generated text to build diverse modality contexts. The structured template is shown in \cref{tab: mc_template}, with human-made physical attributes, corresponding color range, and the mapping between value and color. We assemble the modality context with the color and value information from the image synthesis pipeline. However, such structured text could be meaningless as it is blind to the content in the image, so we utilize the annotation from the source RGB images to filter out the meaningless expression texts.
For the LMM-generated contexts, we merge the generation of modality context with the generation of modality-aware VQA tasks in practice to ensure the LMM won't generate useless modality contexts. We use a 72B Qwen-2.5-VL~\cite{qwen25vl} LMM to complete all multimodal generation and a 32B Qwen-3~\cite{qwen3vl} for language-only generation. The \cref{tab: mc_gen_example} shows two examples of LMM-generated modality context. Detailed generation prompts are listed in \cref{tab: aware_qa_prompt}.


\begin{table}[H]
    \centering
    \caption{Generated Modality Context Examples}
    \label{tab: mc_gen_example}
    \vspace{-0.1in}
    \begin{tcolorbox}[title={\textbf{Generated Modality Context Examples}}]
    \selectfont\ttfamily
    Example 1:\\
    The special color filter in the image represents varying temperature levels. Darker shades indicate colder temperatures, while lighter shades represent warmer temperatures. This allows us to infer how the temperature distribution affects the objects and environment within the bathroom.
    \tcblower
    \selectfont\ttfamily
    Example 2:\\
    In this image, the color filter emphasizes emotional states: blue represents calmness and serenity, while red signifies urgency and alertness. The girl's blue attire and the blue tones around her suggest she is in a state of calm, whereas the red patterns on the blanket indicate areas of potential concern or urgency within the hospital environment.
    \end{tcolorbox}
\end{table}

\section{Training Task Details}
\subsection{Modality-unaware Description Tasks}
The modality-unaware tasks have description tasks and VQA tasks. The generation prompt of description tasks is shown in \cref{tab: sc_prompt} and \cref{tab: pc_prompt} for semantic descriptions and photographic descriptions, respectively. The prompt of semantic description in \cref{tab: sc_prompt} is designed to tell the LMM to extract the shared semantics of the RGB image and the non-RGB image and generate a semantic description from the original RGB descriptions. 
Instead, the prompt of photographic descriptions in \cref{tab: pc_prompt} is designed to capture the unique visual clues from the non-RGB image and generate a detailed description covering the basic visual elements in the image.

\begin{table}[H]
    \centering
    \caption{Semantic Description Generation Prompt Template}
    \label{tab: sc_prompt}
    \vspace{-0.2in}
    \begin{tcolorbox}[title={\textbf{Semantic Description Generation Prompt Template}}]
    \selectfont\ttfamily
    You are an expert at understanding images and describe the scene in the image.\\
    You are given two images: one regular and one non-regular, both are shot from same scene, your task is to describe the scene behind two images.\\
    Regular Image:\\
    <image>\\
    Non-Regular Image:\\
    <image>\\
    Reference Description:\\
    "<caption>"\\
    Your task is to:\\
    1. Analyze the two images, ignore the differences between them, focus on the same scene behind them. Make sure you understand the scene fully.\\
    2. Generate a description that covers the scene. You can check reference description if you are not sure about the scene, but DO NOT copy it.\\
    3. Note that you are describing the scene not the two images, so do not describe any features that affected by the camera or photographic methods.\\
    4. Generate a diverse question that require a subject to describe the scene from a image, please note the subject only can see one image.\\
    5. Output the scene description and the question in format of JSON.\\
    \lbrack REQUIREMENT\rbrack\\
    1. Please ensure that the description is consistent with both images and DO NOT add any features that not exists in the images to the description.\\
    2. The description should focus on the scene in the images, avoid describing details like colors and texture. Also DO NOT describe the features that not showing in both images.\\
    3. Give your ouput in a format of JSON, which has the same structure like this:...
    \tcblower
    \selectfont\ttfamily
    <image>: RGB image and Fabricated image \\
    <caption>: RGB image description
    \end{tcolorbox}
\end{table}
\begin{table}[H]
    \centering
    \caption{Photographic Description Generation Prompt Template}
    \label{tab: pc_prompt}
    \vspace{-0.2in}
    \begin{tcolorbox}[title={\textbf{Photographic Description Generation Prompt Template}}]
    \selectfont\ttfamily
    <image>\\
    You are an expert at understanding image and describe the image. \\
    You are given a non-regular image and a reference description that only describe the scene from a visual assistant, your task is to give a description about the image features that not included in the reference scene description.\\
    Scene Description: \\
    "<caption>" \\
    Your task is to:\\
    1. Analyze the image and scene description, make sure you understand the image fully.\\
    2. Check all elements in the image and write down the features that missing from the scene description. Such features may includes colors, shapes, photographic effects or other features that not involved in the scene description.\\
    3. Based on the missing features, generate a description that includes all missing features but does NOT include described features in the scene description.\\
    4. Generate a diverse question that require a subject to describe the image from the colors, shapes and attributes, etc. Please note the subject only can see the image.\\
    5. Output the missing features, your description and the question in format of JSON.\\
    \lbrack REQUIREMENT\rbrack\\
    1. Please ensure that your description is consistent with the image and DO NOT add any features that not exists in the image to your description.\\
    2. Your description should focus on the missing features in the scene description, especially low-level features like colors and shapes, avoid describing abstract concepts like event and relationship. Also DO NOT describe the features that already included in the scene description.\\
    3. Give your ouput in a format of JSON, which has the same structure like this:...
    \tcblower
    \selectfont\ttfamily
    <image>: Fabricated image.\\
    <caption>: Semantic Description generated before.
    \end{tcolorbox}
    \vspace{-0.3in}
\end{table}

\subsection{Modality-unaware VQA Tasks}
The categories of modality-unaware VQA tasks are listed in \cref{tab: scene_qa} and \cref{tab: photo_qa} with exemplar questions.
As described before, the semantic QA tasks in \cref{tab: scene_qa} are focused on understanding the scene semantics, including object relationships and a broader understanding of the physical world behind the image. And the photographic QA tasks in \cref{tab: photo_qa} are more focused on the visual elements and attributes like number, color and position.
\begin{table}[H] 
    \centering
    \caption{Semantic QA Tasks and Question Templates}
    \label{tab: scene_qa}
    \setlength{\tabcolsep}{12pt}
    \renewcommand{\arraystretch}{1.3}
    \begin{tabularx}{\textwidth}{>{\centering\arraybackslash}m{3.5cm} X}
    \toprule
    Task & \multicolumn{1}{c}{Question Example} \\ 
    \midrule
    Scene Understanding & 
    What are the person and the dog doing?\newline 
    What happened in the scene? \\
    \addlinespace
    Scene Reasoning &
    Why is the man holding a phone?\newline
    Why does the dog have its tone out of its mouth? \newline
    What causes the man to run? \\
    \addlinespace
    Event Extension &
    What will happen if the man drops his phone? \newline
    To solve the problem, which tool does the man need? \newline
    How to prevent the car from crashing? \\
    \addlinespace
    Logical Inference &
    Does the phone belong to the woman? \newline
    What is the most possible relationship between the man and the woman? \newline
    Which person is the bravest one in the scene? \\
    \addlinespace
    Knowledge Inference &
    How does the person feel according to the scene? \newline
    When does this event likely take place? \newline
    What place is it in the image? \\
    \bottomrule
    \end{tabularx}
    \vspace{-0.3in}
\end{table}
\begin{table}[H] 
    \centering
    \caption{Photographic QA Tasks and Question Templates}
    \label{tab: photo_qa}
    \setlength{\tabcolsep}{12pt}
    \renewcommand{\arraystretch}{1.3}
    \begin{tabularx}{\textwidth}{>{\centering\arraybackslash}m{3.5cm} X}
    \toprule
    Task & \multicolumn{1}{c}{Question Example} \\ 
    \midrule
    Color & 
    What is the color of the dog in the image?\newline 
    How many colors can you see on the human? \\
    \addlinespace
    Object Shape &
    What is the shape of the desk? \newline
    Can you describe the shape of the desk? \\
    \addlinespace
    Relative Position &
    What object is next to the person? \newline
    Is the dog to the left of the cat? \newline
    Which side of the desk to the person? \\
    \addlinespace
    Relative Size &
    Which object looks biggest in the image? \newline
    Which object is bigger in the cat and the human? \newline
    Are the two apples on the desk the same size? \\
    \addlinespace
    Object Existence &
    Is there a door in the image? \newline
    Which object does not appear in the image? \newline
    Do all numbers from 0-9 appear in the image? \\
    \addlinespace
    Counting &
    How many objects are there in the image? \newline
    Are there 5 red apples on the desk? \\
    \bottomrule
    \end{tabularx}
\end{table}

The prompt template used to generate these VQA tasks is listed in \cref{tab: unaware_qa_prompt}, which uses the question type and examples to generate corresponding questions and answers. The ``Scene description'' is replaced with semantic descriptions or photographic descriptions when generating VQA separately. The prompt ensures the output is organized in a JSON format and easily converted to training and evaluation data.

\begin{table}[H]
\centering
\caption{Modality-unaware QA generation prompt template}
\label{tab: unaware_qa_prompt}
\vspace{-0.1in}
\begin{tcolorbox}[
title={\textbf{Modality-unaware QA Generation Prompt Template}}]
\begin{Verbatim}[breaklines,breakindent=0em,breaksymbol=]
<image>
You are an expert at understanding images and generating relevant questions and answers based on them to test the ability of a visual assistant.
You are given a non-regular image with a description and a question type with several examples of questions:
Scene description: {caption}
Question Type: {question_type}
Question Examples:
{question_examples}
Your task is to:
1. Analyze the image, scene description, question type, and question examples thoroughly and make sure you understand the scene in the image and the ability demanded by the question type without copy the example questions.
2. Make a strategy to create challenging questions about the image to meet the requirements as below to test the ability of the subject (a visual assistant).
3. Create one question of the given question type based on the strategy and image. Your question should be clear and meaningful, requiring the subject having a strong ability that connecting common knowledge to the scene to solve the question.
4. Provide the correct answer to your question according to the image. The answer should be clear and distinct.
5. Generate at most {negative_samples_num} different plausible incorrect answers that the subject might give if it is doesn't understand the scene or lacks abilities. Adjust the number of incorrect answers according to your question (i.e. if your question only has 'yes' or 'no' as answers, only generate one incorrect answer).
[REQUIREMENT]
1. Please ensure that the image information is fully utilized, and the answer cannot be inferred from the question alone.
2. Make the length of the incorrect answers similar to the correct answer to prevent that the correct answer is always the longest one.
\end{Verbatim}
\end{tcolorbox}
\end{table}
\begin{table}[H]
\begin{tcolorbox}
\begin{Verbatim}[breaklines,breakindent=0em,breaksymbol=]
3. Create questions according to the question type, but please do not completely copy the content of the example questions. You need to generated all kinds of questions related to the question type and the ability behind the question type.
4. You can generate diverse questions requiring broad common knowledge, but ensure the connection between question and answer to be clear and reasonable. Avoid generating question and answers with a lot of guessing and assuming.
5. Give your ouput in a format of JSON, which has the same structure like this:
[FORMAT] Your output MUST be in JSON format as follows:
{
    "strategy": "[HOW TO MAKE IT CHALLENGING]",
    "question": "[YOUR QUESTION]",
    "answer": "[YOUR CORRECT ANSWER]",
    "ability_needed": "[ABILITY TESTED OF YOUR QUESTION TYPE]",
    "incorrect_answers": [
        "[INCORRECT ANSWER 1]",
        "[INCORRECT ANSWER 2]",
        "[INCORRECT ANSWER 3]", ...
    ]
}
\end{Verbatim}
\end{tcolorbox}
\end{table}

\subsection{Modality-aware Description Tasks}
The modality-aware training also contains description tasks and VQA tasks. The prompt used to generate modality descriptions is listed in \cref{tab: mc_gen_prompt}. 
Note that this is a language-only task, for which we use a 32B Qwen-3 language model to generate modality descriptions by combining semantic description, photographic description and modality contexts. The modality contexts are generated along with modality QA tasks to ensure their usefulness. Other two descriptions are generated in the modality-unaware tasks before.
\begin{table}[htbp]
    \centering
    \caption{Modality Description Generation Prompt Template}
    \label{tab: mc_gen_prompt}
    \vspace{-0.1in}
    \begin{tcolorbox}[title={\textbf{Modality Description Generation Prompt Template}}]
    \selectfont\ttfamily
    You are an expert at understanding image descriptions and rewrite them to a better description.\\
    You are given tow description about a special colored image, one is focusing on describe the scene and the other one is focusing on describe the detailed colors and shapes:\\
    Scene description:\\
    <semantic\_caption>\\
    Photographic description:\\
    <photographic\_caption>\\
    Modality description:\\
    <modality\_context>\\
    Your task is to:\\
    1. Analyze the tow descriptions, make sure you understand the image according to the description.\\
    2. Analyze the modality description which explained what the color filter is trying to emphasize.\\
    3. According to the modality description, combine the scene description and photographic description to a more completed description that cover all features in the image combined with the modality description.\\
    4. Create one question that asking a subject describe the image based on the modality description and image. You can assume the subject only can see the image and modality description.\\
    \lbrack REQUIREMENT\rbrack
    1. Please ensure that the image information and the modality description are fully utilized in the completed description. Every details in the image is described in the answer with the modality description.\\
    2. Avoid generating question with a lot of guessing and assuming. However you can make up some extra knowledge in the modality description, it will be provided to subject.\\
    3. Give your ouput in a format of JSON, which has the same structure like this:\\
    \tcblower
    \selectfont\ttfamily
    <semantic\_caption>: Semantic Description generated before.\\
    <photographic\_caption>: Photographic Description generated before.\\
    <modality\_context>: Modality Context fabricated before.\\
    \end{tcolorbox}
\end{table}

\subsection{Modality-aware VQA Tasks}
The sub-tasks and their question examples are listed in \cref{tab: modality_qa}. They are generated using the prompt listed in \cref{tab: aware_qa_prompt} along with the modality context. We use a similar generation format to ensure the usability of the LMMs' output. Note that the ``modality description'' in \cref{tab: aware_qa_prompt} actually refers to modality context, as it is asking the model to generate a piece of text describing the mapping between color patterns and physical meanings.
\begin{table}[H]
    \centering
    \caption{Modality QA Tasks and Question examples}
    \label{tab: modality_qa}
    \setlength{\tabcolsep}{12pt}
    \renewcommand{\arraystretch}{1.3}
    \begin{tabularx}{\textwidth}{>{\centering\arraybackslash}m{3.5cm} X}
    \toprule
    Task & \multicolumn{1}{c}{Question Example} \\ 
    \midrule
    Modality Understanding & 
    Based on the color meaning, what are the person and the dog doing? \newline 
    What happened in the scene shown by the special color filter? \\
    \addlinespace
    Modality Reasoning &
    Why does the dog look red by the color filter? \newline
    Why does the human look irregularly under the color patterns? \newline
    What causes the man to run? \\
    \addlinespace
    Event Extension &
    What will happen if the man drops his phone, considering the ground is red? \newline
    To solve the problem, which tool does the man need if the color red represents hardness? \newline
    Which person should be quarantined if the color means temperature? \\
    \addlinespace
    Logical Inference &
    Is the car running? \newline
    Which are the most possible tools to open the door? \newline
    Which person is the happiest one in the scene based on the temperature shown by the color? \\
    \addlinespace
    Knowledge Inference &
    How does the person feel according to the scene if the color represents blood flow? \newline
    When does this event likely take place, considering the ultraviolet level? \newline
    What place is it in the image if the color indicates the gravity level? \\
    \bottomrule
    \end{tabularx}
\end{table}
\begin{table}[H]
\centering
\caption{Modality-aware QA generation prompt template}
\label{tab: aware_qa_prompt}
\vspace{-0.1in}
\begin{tcolorbox}[
title={\textbf{Modality-aware QA Generation Prompt Template}}]
\begin{Verbatim}[breaklines,breakindent=0em,breaksymbol=]
<image>
You are an expert at understanding images and generating relevant questions and answers based on them to test the ability of a visual assistant.
You are given a non-regular image that has special color filter and a question type with several examples of questions:
Scene description: {caption}
Question Type: {question_type}
Question Examples:
{question_examples}
Your task is to:
1. Analyze the image, scene description, question type, and question examples thoroughly and make sure you understand the scene in the image and the ability demanded by the question type without copy the example questions.
2. Analyze the special color pattern of the image and generate a modality description to explain what the color filter is trying to emphasize.
3. Make a strategy to create challenging questions about the image and its color filter to meet the requirements as below to test the ability of the subject (a visual assistant).
4. Create one question of the given question type based on the strategy and image according to the modality description. Your question should be clear and meaningful, requiring the subject having a strong ability that connecting common knowledge to the modality description to solve the question.
5. Provide the correct answer to your question according to the image and the modality description. The answer should be clear and distinct.
6. Generate at most {negative_samples_num} different plausible incorrect answers that the subject might give if it is doesn't understand the the modality description or lacks abilities. Adjust the number of incorrect answers according to your question (i.e. if your question only has 'yes' or 'no' as answers, only generate one incorrect answer).
[REQUIREMENT]
1. Please ensure that the image information and the modality description are fully utilized, and make sure the answer cannot be inferred from the question alone.
2. Make the length of the incorrect answers similar to the correct answer to prevent that the correct answer is always the longest one.
3. Create diverse questions according to the question type, but please DO NOT completely copy the content of the example questions. You need to generated all kinds of questions related to the question type and the ability behind the question type.
\end{Verbatim}
\end{tcolorbox}
\end{table}
\begin{table}[H]
\begin{tcolorbox}
\begin{Verbatim}[breaklines,breakindent=0em,breaksymbol=]
4. You can generate diverse questions requiring broad common knowledge, but ensure the connection between question and answer to be clear and reasonable. Avoid generating question and answers with a lot of guessing and assuming. However you can make up some extra knowledge in the modality description, it will be provided to subject.
5. Give your ouput in a format of JSON, which has the same structure like this:
[FORMAT] Your output MUST be in JSON format as follows:
{
    "strategy": "[HOW TO MAKE IT CHALLENGING]",
    "modality_description": "[MAKE UP SOME MEANINGS FOR THE COLOR PATTERN]",
    "question": "[YOUR QUESTION]",
    "answer": "[YOUR CORRECT ANSWER]",
    "ability_needed": "[ABILITY TESTED OF YOUR QUESTION TYPE]",
    "incorrect_answers": [
        "[INCORRECT ANSWER 1]",
        "[INCORRECT ANSWER 2]",
        "[INCORRECT ANSWER 3]", ...
    ]
}
\end{Verbatim}
\end{tcolorbox}
\end{table}

\end{document}